\documentclass[letterpaper, 10 pt, conference]{ieeeconf}

\IEEEoverridecommandlockouts
\PassOptionsToPackage{numbers,sort&compress}{natbib}

\usepackage{xcolor}
\makeatletter
\let\NAT@parse\undefined
\makeatother
\usepackage{natbib}

\usepackage{amsmath,amssymb}
\usepackage{graphicx}
\IfFileExists{latexml.sty}{\usepackage{latexml}}{\newif\iflatexml}
\usepackage{booktabs}
\usepackage{multirow}
\usepackage{float}
\usepackage{placeins}
\usepackage{tikz}
\usepackage{xspace}
\usepackage{pgfplots}
\usepackage[hidelinks]{hyperref}
\usetikzlibrary{arrows.meta,positioning,calc}
\pgfplotsset{compat=1.18}
\definecolor{i3lorange}{RGB}{196,91,40}

\makeatletter
\let\iil@templatecaption\@makecaption
\long\def\@makecaption#1#2{%
\ifx\@captype\@IEEEtablestring
\parbox[t]{\hsize}{\footnotesize\noindent #1:~#2}%
\@IEEEtablecaptionsepspace
\else
\iil@templatecaption{#1}{#2}%
\fi}
\makeatother

\newcommand{\algo}{I3L\xspace}

\title{\LARGE \bf
I3L: Iterative Interactive Imitation Learning\\for Whole-Body Mobile Manipulation
}

\author{Anonymous Author(s)}

\author{Wensi Ai$^{1,*}$, Christopher Luey$^{1,*}$, Karthik Dharmarajan$^{1,*}$, Evans Han$^{1}$, Yunfan Jiang$^{1}$, Stef Ren$^{1}$, \\ Jianwen Xie$^{2}$, Jiajun Wu$^{1}$, Li Fei-Fei$^{1,\dagger}$, Ruohan Zhang$^{1,3,\dagger}$ \\ $^{1}$Stanford University \quad $^{2}$Lambda Inc. \quad $^{3}$Northwestern University\thanks{$^{*}$Equal contribution. $^{\dagger}$Equal advising. Correspondence: Wensi Ai (wsai@stanford.edu).}}

\begin{document}
\iflatexml
\maketitle
\begin{figure*}
\centering
\includegraphics[width=\textwidth]{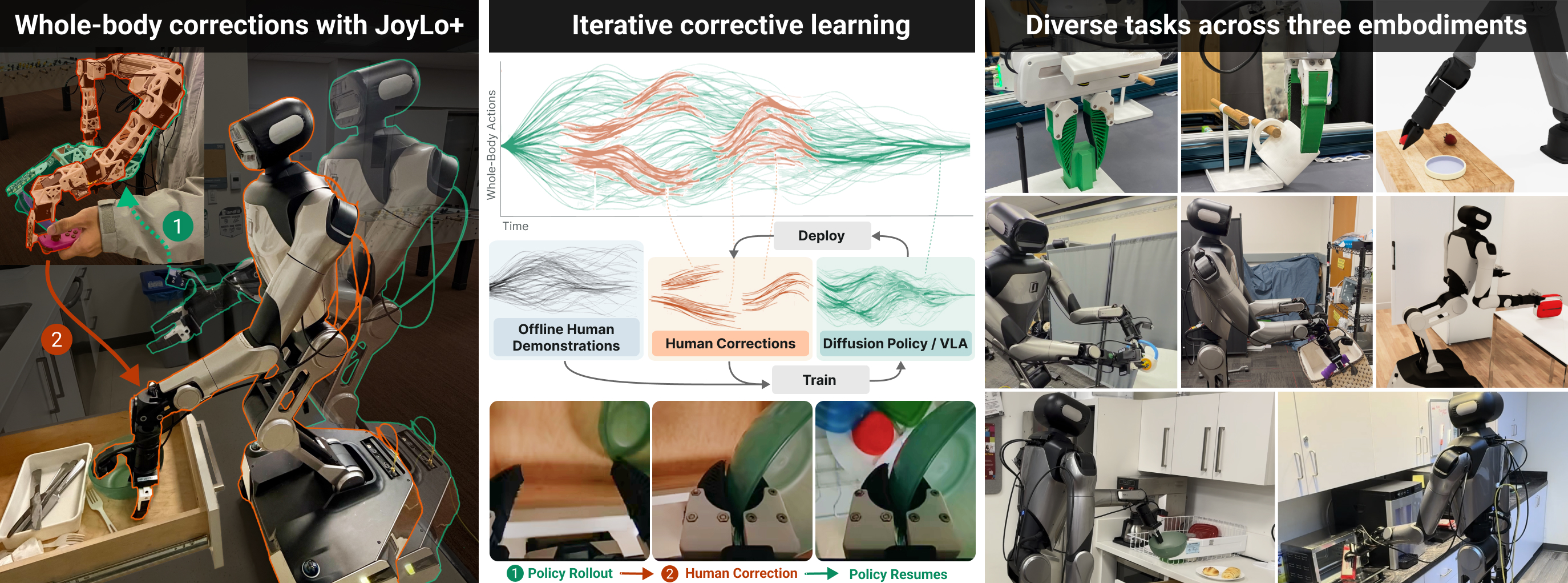}
\caption{\textbf{I3L combines interactive correction with iterative policy refinement.} Left: 1) the policy approaches a failure, prompting the operator to take over with JoyLo+ and provide targeted corrections, 2) the operator uses JoyLo+ to take over during autonomous execution. Center: corrections are aggregated with prior demonstrations for policy retraining and redeployment. Right: eight evaluation tasks across three embodiments.}
\label{fig:pull}
\end{figure*}
\else
\newsavebox{\titlefigure}
\makeatletter
\sbox{\titlefigure}{%
\begin{minipage}{\textwidth}
\def\@captype{figure}
\centering
\includegraphics[width=1.0\textwidth]{figures/pull2.jpg}
\vspace{-2em}
\caption{\textbf{I3L combines interactive correction with iterative policy refinement.} Left: 1) the policy approaches a failure, prompting the operator to take over with JoyLo+ and provide targeted corrections, 2) the operator uses JoyLo+ to take over during autonomous execution. Center: corrections are aggregated with prior demonstrations for policy retraining and redeployment. Right: eight evaluation tasks across three embodiments.}

\label{fig:pull}
\end{minipage}%
\vspace{-3em}
}
\g@addto@macro\@maketitle{\par\medskip\noindent\usebox{\titlefigure}\par}
\makeatother
\maketitle
\fi
\thispagestyle{empty}
\pagestyle{empty}
\raggedbottom
\setlength{\textfloatsep}{10pt plus 2pt minus 2pt}

\begin{abstract}
Policies trained offline from teleoperated demonstrations can degrade outside their training distribution. This gap is particularly acute in long-horizon whole-body mobile manipulation, where long horizons amplify compounding errors and coordinated arm, torso, and base motion makes exhaustive demonstration coverage impractical. We introduce I3L, an integrated hardware and learning system that combines whole-body correction with iterative policy refinement to address this last-mile reliability gap. Its low-cost, open-source JoyLo+ interface uses policy-tracking leader arms and touch-triggered takeover to enable coordinated corrections across the body parts, while iterative deployment and retraining allow operators to target failures of the updated policy. We evaluate I3L across three robot embodiments and eight simulated and real-world tasks, including four long-horizon whole-body mobile manipulation tasks. Controlled simulation experiments show that four-round HG-DAgger achieves the highest final success on both tasks. Under this protocol, I3L improves autonomous success over the base policies by an average of 68.3 percentage points and achieves at least 90\% success on every task, outperforming demonstration-only training by 35.6 percentage points on average with the same number of training trajectories. Averaged across tasks, takeover duration per episode decreases by 72\% over four rounds. Together, these results show that I3L helps close the last-mile gap between offline training and reliable deployment and provides practical insights into collecting and using corrections for long-horizon whole-body mobile manipulation. Videos, visualizations, and appendix are at \url{https://i-3l.github.io/}.

\end{abstract}

\section{Introduction}
\label{sec:intro}

Visuomotor imitation learning has advanced through improved action representations and policy learning methods~\citep{chi2023diffusion,zhao2023learning}, alongside generalist robot policies trained on increasingly diverse datasets~\citep{brohan2023rt2,oxe2023,khazatsky2024droid,kim2024openvla,octo2024,black2024pi0,intelligence2025pi05,bjorck2025gr00t}. These developments enable language-conditioned manipulation, cross-embodiment transfer, and long-horizon mobile manipulation. Yet when deploying on robots to finish a real task, they still fail in frustrating ways. Many policies use \emph{offline imitation learning} from a fixed set of demonstrations, leaving them vulnerable to distribution shift and compounding errors during actual deployment~\citep{ross2011dagger}.

This covariate-shift problem is particularly prevalent in whole-body mobile manipulation, which couples over twenty actuated degrees of freedom with an omnidirectional base across varied scenes and objects. Even when a policy performs most stages competently, failures at a few critical stages on a single body part can prevent full-task completion. Closing this \emph{last-mile gap} therefore depends not just on collecting more data, but on where it is collected: targeted corrections at bottleneck states can be more valuable than demonstrations of already-mastered behavior.

These challenges motivate supervision that directly addresses deployment-time errors. \emph{Interactive imitation learning} (IIL) lets a human supervise the policy at deployment and correct it where it fails, so the next iteration trains on the policy's own induced distribution~\citep{celemin2022interactive,ross2011dagger,kelly2019hgdagger,mandlekar2020iwr,liu2023sirius,jiang2024transic,wu2025robocopilot,xu2025crdagger}. Targeted intervention concentrates teleoperation at failure-prone states, reducing the need for end-to-end human control. Prior work has demonstrated repeated corrective learning, compliant intervention interfaces, and mobile bimanual manipulation~\citep{liu2023sirius,wu2025robocopilot,xu2025crdagger}. However, systematic evidence remains limited for corrective learning across diverse long-horizon tasks requiring coordinated base, torso, and arm motion. Making such interventions practical requires an interface that lets the operator take control promptly while remaining aligned with autonomous execution. Effective corrective learning also requires choosing how to train on corrections and whether to collect them against a fixed or updated policy.

To this end, we present \textbf{\algo} (Iterative Interactive Imitation Learning), an integrated hardware and learning system for iterative whole-body correction in long-horizon mobile manipulation. Its JoyLo+ interface combines policy-tracking leader arms with capacitive-touch-triggered switching to support smooth takeover, while repeated deployment and retraining allow collection to target failures of the updated policy. Using this system, we evaluate iterative whole-body correction on four long-horizon real-world BEHAVIOR tasks~\citep{li2024behavior1k}, complemented by controlled comparisons of collection schedules and learning strategies and analysis of intervention burden across simulation and real-world tasks. Our contributions are:

\textbf{1. An integrated system for iterative whole-body correction.} \algo combines JoyLo+, our whole-body correction interface, with repeated policy deployment and retraining for long-horizon mobile manipulation. On all four real-world BEHAVIOR tasks~\citep{li2024behavior1k}, \algo helps close the last-mile reliability gap by achieving over 90\% success, outperforming demonstration-only training by 30 percentage points on average under the same trajectory budgets.

\textbf{2. JoyLo+: a touch-triggered whole-body correction interface.} Extending GELLO~\citep{wu2023gello} and JoyLo~\citep{jiang2025brs}, our low-cost, open-source interface combines capacitive-touch detection with position/current-mode switching on motorized leader arms, enabling global takeover for coordinated arm, torso, and base corrections.

\textbf{3. Studies of corrective learning and operator intervention.} Under the same trajectory budgets, we compare correction-learning methods on two simulation tasks. Four-round HG-DAgger~\citep{kelly2019hgdagger} achieves the highest final mean success on both tasks. A user study further shows declining cumulative intervention share across operators with varied teleoperation experience.

\begin{figure*}[!t]
\centering
\includegraphics[width=\textwidth]{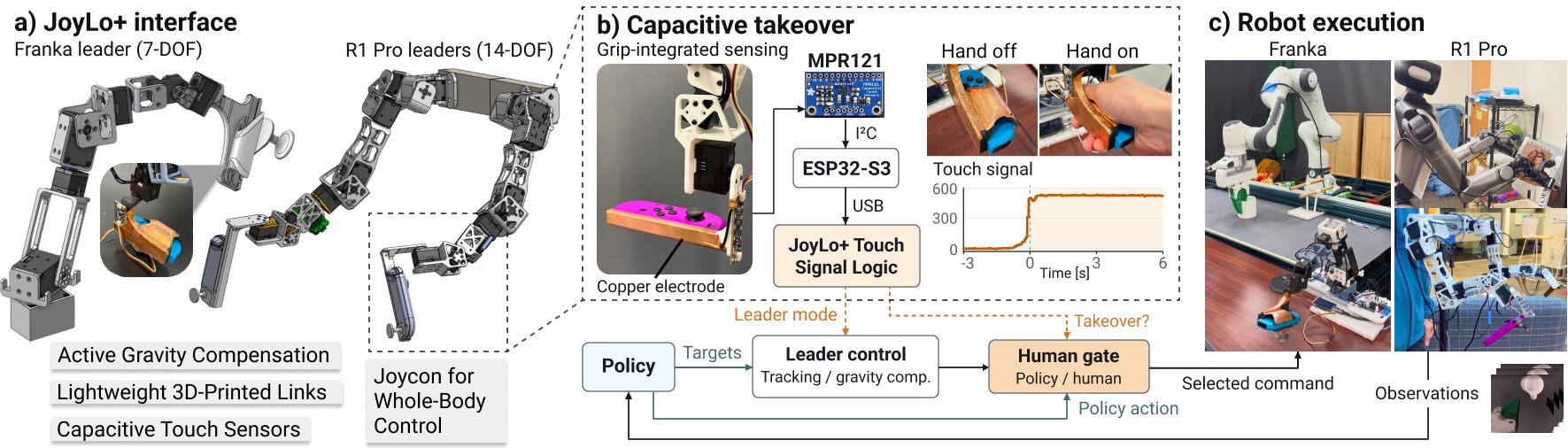}
\vspace{-1.5em}
\caption{\textbf{JoyLo+ correction interface.} Motorized arm leaders track policy targets during autonomous execution and switch to compliant current control during operator takeover. Capacitive touch sensors trigger takeover, while JoyCon inputs control the torso, mobile base, and grippers. The interface supports both single-arm and dual-arm configurations.}
\label{fig:system}
\vspace{-1.5em}
\end{figure*}

\section{Related Work}
\label{sec:related}

\textbf{Data-collection Interfaces for Correction.} Teleoperation interfaces determine how easily and precisely operators can collect data and provide corrections~\citep{cheng2024television}. Existing data-collection approaches use 3D mice~\citep{jiang2024transic}, smartphones~\citep{mandlekar2020iwr,mandlekar2018roboturk}, motion capture~\citep{wang2024dexcap}, vision-based arm--hand tracking~\citep{qin2023anyteleop}, handheld and wearable demonstration interfaces~\citep{chi2024umi,xu2026hommi,ha2024umilegs,xu2025dexumi,yang2024ace}, and kinematic-twin interfaces~\citep{zhao2023learning,fu2024mobile,wu2023gello,jiang2025brs}. TeleMoMa~\citep{dass2024telemoma} combines heterogeneous inputs for whole-body mobile teleoperation, while FACTR provides bilateral force-feedback teleoperation for contact-rich tasks~\citep{liu2025factr}. Correcting an executing policy additionally requires smooth transitions between autonomous control and human intervention. RoboCopilot and HACTS synchronize their leaders with the robot during autonomous execution to support takeover~\citep{wu2025robocopilot,xu2025hacts}; CR-DAgger supports kinesthetic corrections through a force-instrumented handle~\citep{xu2025crdagger}. Human-Robot Copilot uses motion scaling for fine-grained corrective teleoperation~\citep{yan2026copilot}, and DexHiL integrates coordinated arm-hand intervention for dexterous VLA post-training~\citep{han2026dexhil}. I3L's JoyLo+ combines policy-tracking leaders and touch-triggered takeover to let operators switch from autonomous execution to coordinated correction of the arms, torso, and base.

\textbf{Interactive Imitation Learning Algorithms.} DAgger addresses covariate shift by collecting expert labels on policy-visited states~\citep{ross2011dagger}. DART instead injects noise during demonstration collection to expose recovery states~\citep{laskey2017dart}. Subsequent work studies when to request human intervention and how to use the resulting data. HG-DAgger lets humans gate intervention~\citep{kelly2019hgdagger}, an approach also used in BC-Z~\citep{jang2021bcz}; SafeDAgger reduces expert queries with a learned safety classifier~\citep{zhang2017safedagger}, while ThriftyDAgger gates queries using novelty and risk estimates~\citep{hoque2022thrifty}. LazyDAgger reduces control switches to limit supervisor burden~\citep{hoque2021lazydagger}. Sirius-Fleet extends corrective learning across tasks with world-model-based anomaly prediction~\citep{liu2024siriusfleet}. Methods also differ in how they use deployment data: IWR balances intervention and non-intervention samples~\citep{mandlekar2020iwr}, SIRIUS reweights data using estimated human trust~\citep{liu2023sirius}, and DexHiL prioritizes corrective segments~\citep{han2026dexhil}. TRANSIC and CR-DAgger learn residual corrections~\citep{jiang2024transic,xu2025crdagger}, while FlowCorrect adapts flow policies from sparse relative corrections with a frozen backbone~\citep{welte2026flowcorrect}. Related reinforcement-learning approaches also incorporate human corrections, including RECAP~\citep{intelligence2025recap} and HIL-SERL~\citep{luo2024hilserl}. I3L examines how correction-learning strategies and collection schedules affect policy success and intervention burden over successive rounds under the same trajectory budgets.

\textbf{Whole-Body Mobile Manipulation.} Whole-body mobile manipulation has advanced through demonstration-based systems and generalist policies~\citep{fu2024mobile,fu2024humanplus,jiang2025brs,intelligence2025pi05,bjorck2025gr00t,wu2024tidybot,ha2024umilegs,xu2026hommi,he2024omnih2o}. Building on these capabilities, recent work has explored corrective learning during deployment. RoboCopilot evaluates corrective learning on mobile part transport and a separate four-stage kitchen task~\citep{wu2025robocopilot}, while XR-2 reports iterative DAgger results for clothes folding~\citep{xu2026scaling}. HITL-TAMP achieves high success on long-horizon contact-rich tasks by integrating learned manipulation skills within a task-and-motion-planning framework~\citep{mandlekar2023hitltamp}. Crucially, systematic evidence remains limited on whether iterative correction can close the last-mile reliability gap and achieve consistently high success across diverse long-horizon whole-body mobile manipulation tasks. I3L addresses this question across a suite of long-horizon mobile household tasks, while tracking how intervention burden changes across subtasks and body parts.

\section{Method}
\label{sec:method}

\algo integrates JoyLo+ correction hardware with iterative policy learning. JoyLo+ lets operators follow the policy's motion, take over at failure states, and provide targeted corrections. The system records these corrections with intervention labels, aggregates them with prior demonstrations, and updates the policy for the next deployment round.

\begin{table}[H]
\centering
\caption{Teleoperation and correction interfaces. $\checkmark$: demonstrated; --: not demonstrated in the cited work.}
\label{tab:hardware-capabilities}
\footnotesize
\newcommand{\hwYes}{\ensuremath{\checkmark}}
\newcommand{\hwNo}{\textcolor{black!55}{--}}
\newcommand{\hwOurs}{\textcolor{i3lorange}{\ensuremath{\checkmark}}}
\newcommand{\hwHead}[1]{{\scriptsize #1}}
\setlength{\tabcolsep}{0.8pt}
\renewcommand{\arraystretch}{1.1}
\begin{tabular*}{\columnwidth}{@{\extracolsep{\fill}}lccccc@{}}
\toprule
\hwHead{\textbf{System}} & \hwHead{Mobile} & \hwHead{Bimanual} & \hwHead{Multi-embodiment} & \hwHead{Actuated} & \hwHead{Correction} \\
\midrule
ALOHA/2~\citep{zhao2023learning,aloha2team2024} & \hwNo & \hwYes & \hwNo & \hwNo & \hwNo \\
GELLO~\citep{wu2023gello} & \hwNo & \hwYes & \hwYes & \hwNo & \hwNo \\
Mobile ALOHA~\citep{fu2024mobile} & \hwYes & \hwYes & \hwNo & \hwNo & \hwNo \\
TeleMoMa~\citep{dass2024telemoma} & \hwYes & \hwYes & \hwYes & \hwNo & \hwNo \\
JoyLo~\citep{jiang2025brs} & \hwYes & \hwYes & \hwNo & \hwYes & \hwNo \\
UME~\citep{liang2026ume} & \hwYes & \hwYes & \hwYes & \hwYes & \hwNo \\
\addlinespace[2pt]
CR-DAgger~\citep{xu2025crdagger} & \hwNo & \hwNo & \hwNo & \hwNo & \hwYes \\
HACTS~\citep{xu2025hacts} & \hwNo & \hwNo & \hwNo & \hwYes & \hwYes \\
RoboCopilot~\citep{wu2025robocopilot} & \hwYes & \hwYes & \hwNo & \hwYes & \hwYes \\
\midrule
\textbf{JoyLo+ (ours)} & \hwOurs & \hwOurs & \hwOurs & \hwOurs & \hwOurs \\
\bottomrule
\end{tabular*}
\par\smallskip
\parbox{\columnwidth}{\footnotesize
\emph{Mobile}: coordinated arm and mobile-base control; torso where available.
\emph{Multi-embodiment}: distinct arm kinematics on real hardware.
\emph{Actuated}: leader motors respond to robot state, torque, or policy targets.
\emph{Correction}: takeover or kinesthetic guidance during autonomous rollouts.
JoyLo+: R1~Pro (mobile, bimanual); R1~Pro and Franka (multi-embodiment).}
\end{table}

\subsection{JoyLo+: A Hardware Interface for Whole-Body Policy Intervention}
\label{sec:takeover}
\label{sec:hardware}

JoyLo+ is an actuated teleoperation interface for data collection and intervention during autonomous policy execution. We implement \textbf{R1~Pro JoyLo+}, with two seven-joint leaders for bimanual whole-body mobile manipulation, and \textbf{Franka JoyLo+}, with one seven-joint leader for single-arm table-top manipulation. Building on GELLO's kinematic leaders~\citep{wu2023gello} and JoyLo's interface~\citep{jiang2025brs}, JoyLo+ combines policy-target tracking, capacitive-touch detection, and compliant hand guidance (Figure~\ref{fig:system}). Table~\ref{tab:hardware-capabilities} compares it with existing interfaces, highlighting the capabilities enabled by the hardware and control design described below:

\paragraph{Actuated leaders for tracking and correction}
Each configuration uses 3D-printed links, joint mappings, and calibration matched to its robot. During policy rollouts, the actuated leaders track the policy's arm joint targets, keeping JoyLo+ aligned with the commanded action for smooth takeover. During intervention, gravity compensation supports the leaders while the operator guides their motion. The component cost for one R1~Pro leader assembly and the listed accessories totals \$1,257.49. A detailed cost breakdown is available in the appendix and on our project website.

\begin{figure*}[!t]
\centering
\includegraphics[width=\textwidth]{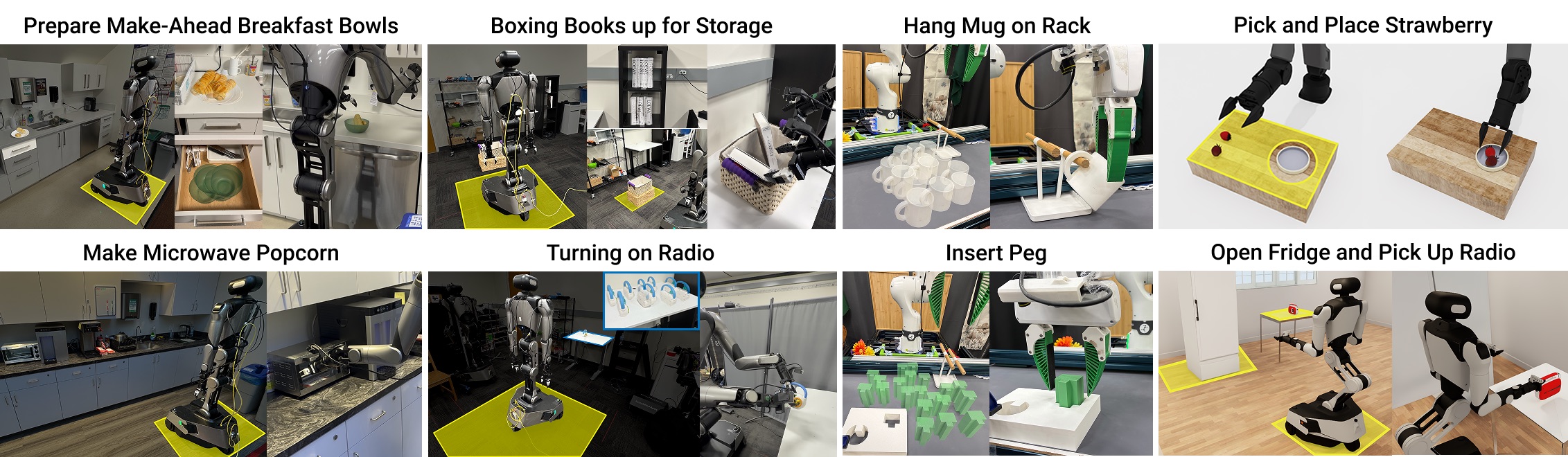}
\vspace{-2em}
\caption{\textbf{Evaluation tasks and initialization randomization.} Representative images show initial configurations and successful task outcomes. The eight tasks span 29 annotated subtasks (2--6 per task). Across the four real-world R1~Pro tasks, task executions average approximately 83 seconds, with book storage and breakfast preparation averaging close to two minutes each. Yellow mats indicate the R1~Pro base initialization regions as well as where some task-relevant objects are randomized at the start of each episode.}
\label{fig:tasks}
\vspace{-1.5em}
\end{figure*}

\paragraph{Grip-integrated sensing for direct takeover}
JoyLo+ uses grip-integrated copper electrodes, an MPR121 capacitive-sensing board, and an ESP32-S3 controller to report touch state to the computer. This enables instant touch-triggered takeover without requiring a separate button press or movement of the leader arm. Different thresholds for detecting touch and release, together with a short release delay, prevent brief losses of contact from unintentionally returning control to the policy.

\paragraph{Coordinated takeover and policy resumption}
On R1~Pro, touching a grip or providing a JoyCon (with joysticks and buttons) input transfers command of both arms, the torso, the base, and the grippers together. The leaders guide the arms, while JoyCons control the torso, base, and grippers. This ensures operators have full control over the robot and prevents autonomous commands from interfering with the correction. In both configurations, releasing the interface returns control to the policy, which replans from the corrected state as the leaders resume tracking.

\subsection{Intervention Detection and Recording}
\label{sec:detection}

JoyLo+ uses the takeover state to switch control and label corrective experience. At each timestep, we record the intervention-state label $s_t$, observation $\mathbf{o}_t$, executed robot command $\mathbf{a}^{\mathrm{exec}}_t$, and policy-proposed action $\mathbf{a}^{\mathrm{\pi}}_t$. $\mathbf{a}^{\mathrm{exec}}_t$ records the operator’s action during takeover and the policy’s action during autonomous rollout. Each frame is assigned a label $s_t$: original demonstration ($1$), active intervention ($2$), the 15-frame pre-intervention window ($3$), or autonomous execution ($4$). This enables us to compare how different learning strategies use the collected data (Section~\ref{sec:q4}).

\subsection{Iterative Interactive Imitation Learning}
\label{sec:protocol}

\paragraph{Data aggregation}
Starting from a demonstration dataset $\mathcal{D}_0$, round $r$ deploys the previous policy and collects a labeled rollout dataset $\mathcal{R}_r$ through JoyLo+. Under HG-DAgger filtering~\citep{kelly2019hgdagger}, which achieves the best results in the controlled experiment (Section~\ref{sec:q4}), the training dataset becomes
\begin{equation}
    \mathcal{D}_r = \mathcal{D}_{r-1} \cup
    \{(\mathbf{o}_t,\mathbf{a}^{\mathrm{exec}}_t, \mathbf{a}^{\mathrm{\pi}}_t)\in\mathcal{R}_r : s_t=2\}.
\end{equation}
Thus, original demonstrations and active corrections from all rounds provide the training data. For DP action-chunk training, filtering applies to chunk starts; contiguous targets may extend across takeover transitions (Appendix~\ref{app:processing}). Policy architectures and training settings are specified in Section~\ref{sec:setup}.

\paragraph{Policy retraining and redeployment}
After each round, we retrain the policy on the aggregated dataset $\mathcal{D}_r$. Training starts from the same pretrained checkpoint for the VLA and the same initial weights for diffusion policies, as using a common initialization keeps the training starting point consistent across rounds as the dataset grows. The updated policy is then deployed to collect the next round of corrections. This iterative process allows operators to target the policy’s remaining failures after each round of retraining.

\section{Experiments}
\label{sec:experiments}

I3L's primary application is real-world whole-body mobile manipulation, which we showcase on R1~Pro tasks. Franka tasks examine I3L's capability in contact-rich scenarios requiring precise manipulation, while simulation tasks systematically compare correction collection schedules and learning strategies. We ask: \textbf{Q1}: Do targeted corrections address the last mile problem by improving task success over additional demonstrations? \textbf{Q2}: How does iteration affect policy performance and human effort? \textbf{Q3}: How do whole-body corrections engage the arms, torso, and mobile base? \textbf{Q4}: How do correction-learning strategies compare? \textbf{Q5}: How do policy improvement and intervention trends vary across operators with varying teleoperation experience?

\begin{figure*}[!t]
\centering
\setlength{\abovecaptionskip}{3pt}
\includegraphics[width=\textwidth]{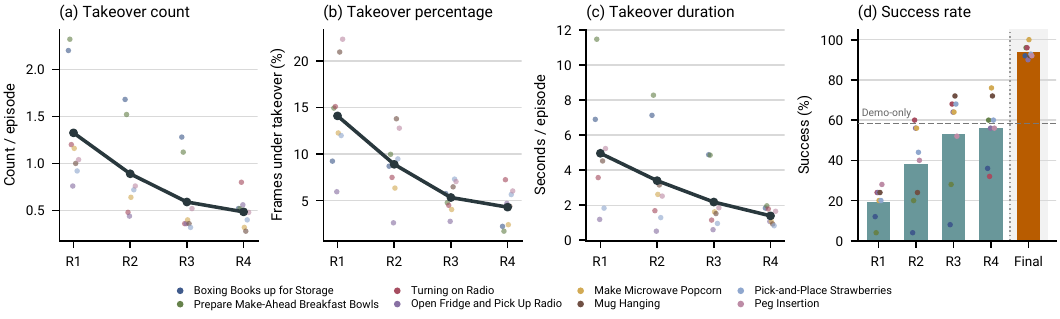}
\caption{\textbf{Intervention burden and task completion across correction rounds.} In (a--c), the dark line shows the equal-weight mean across eight tasks; (b) measures the fraction of execution under intervention, and (c) measures time under intervention per episode. In (d), rounds~1--4 show takeover-free collection episodes. The final bar and dashed demo-only baseline (58.3\%) compare autonomous evaluation success under the same trajectory budgets.}
\label{fig:intervention_trends}
\end{figure*}

\subsection{Experimental Setup}
\label{sec:setup}

Repeated collection, retraining, physical resets, and evaluation make long-horizon mobile-manipulation studies resource-intensive. We therefore concentrate controlled algorithm comparisons in simulation and use real-world experiments to evaluate the selected protocol across diverse  household tasks.

\paragraph{Tasks and platforms}
Evaluation is conducted on eight tasks (with twenty-nine subtasks) across simulated and real-world domains involving three robot embodiments. Figure~\ref{fig:tasks} shows the initial configuration and a successful execution for each task, along with the randomization applied to the task objects and to the robot base pose at the start of every episode. Figure~\ref{fig:corrections} shows the annotated subtask stages and their intervention patterns across four correction rounds.

\textbf{Controlled simulation (two tasks).} A1 Pick-and-Place Strawberries and R1~Pro Open Fridge and Pick Up Radio are done in simulation. They comprise four and three annotated subtasks respectively, with task executions averaging approximately 30 seconds. We set up our simulation environments using the OmniGibson simulator from BEHAVIOR~\citep{li2024behavior1k,ge2024behaviorvisionsuite}. For these two tasks, scripted oracles generate demonstrations and corrective trajectories using cuRobo motion planning~\citep{sundaralingam2023curobo}, enabling efficient multi-round experiments and controlled algorithm comparisons. A human operator decides when to intervene and when to resume policy execution, while the scripted oracle generates the corrective motions. Both tasks are designated for single-round versus multi-round collection and correction-learning comparisons, including residual learning, BC (all data), SIRIUS, and HG-DAgger. This study isolates collection and learning choices under controlled conditions.

\textbf{Fixed-base manipulation (two tasks).} Franka peg insertion and mug hanging demonstrate JoyLo+ in a single-arm configuration and evaluate I3L on precision-oriented, contact-rich tasks. Each task consists of two annotated subtasks on sequential pick and place, with task executions averaging approximately 24 seconds for peg insertion and 23 seconds for mug hanging. Both tasks require precise object alignment during contact: aligning the mug handle with the rack and the peg with its insertion opening.

\textbf{Whole-body mobile manipulation (four tasks).} R1~Pro performs four long-horizon real-world BEHAVIOR tasks~\citep{li2024behavior1k}: Prepare Make-Ahead Breakfast Bowls, Boxing Books up for Storage, Make Microwave Popcorn, and Turning on Radio, consisting of six, five, four, and three annotated subtasks, respectively. These tasks require sequential object interactions and coordinated mobile manipulation. Task executions average approximately 83 seconds across these tasks, with breakfast preparation and book storage averaging about two minutes each.

\paragraph{Evaluation protocol}
During human-supervised rollouts for correction data collection, autonomous completion denotes episodes finished without human takeover. A subtask is counted as intervention-free when no takeover occurs within its annotated interval. All real-world evaluations use 25 rollouts per condition. Simulation evaluations are averaged across five evaluation seeds, with 50 rollouts per seed.

\paragraph{Training}
A1 and Franka use diffusion policies (DP)~\citep{chi2023diffusion} trained on task data only (ImageNet-initialized ResNet-18 encoders~\citep{deng2009imagenet,kaimingresnet}, UNet backbone, DDIM~\citep{song2021ddim}, AdamW~\citep{loshchilov2019adamw}, batch size 128, cosine LR). R1~Pro uses $\pi_{0.5}$~\citep{intelligence2025pi05}, a pretrained vision-language-action (VLA) model. Each round trains on accumulated demonstrations and active corrections, restarting from the same weight initialization for diffusion policies or the same pretrained checkpoint for the VLA. We train all $\pi_{0.5}$ policies for 40k steps and DP for 200k steps before evaluation, and evaluate the last checkpoint for consistency. Appendix~\ref{app:policy} provides detailed training and deployment settings.

\subsection{Q1: Targeted Corrections Improve Task Performance}
\label{sec:q1}

\begin{table}[!htbp]
\centering
\caption{Autonomous full-task success rates before and after additional data collection. Base policies receive 100 additional demonstrations or 100 corrective trajectories through four HG-DAgger rounds. Simulation rates are means across five evaluation seeds.}
\label{tab:task_summary}
\scriptsize
\setlength{\tabcolsep}{2pt}
\resizebox{\columnwidth}{!}{%
\begin{tabular}{@{}lrrr@{}}
\toprule
Task & Base & 100 Demos & \textbf{100 Corr. (Ours)} \\
\midrule
\multicolumn{4}{@{}l}{\textit{Simulation}} \\
Pick-and-Place Strawberries & 53.2\% & 55.6\% & \textbf{92.8\%} \\
Open Fridge and Pick Up Radio & 39.6\% & 30.4\% & \textbf{90.0\%} \\
\midrule
\multicolumn{4}{@{}l}{\textit{Real world (Franka, Diffusion Policy)}} \\
Mug Hanging & 6/25 (24\%) & 17/25 (68\%) & \textbf{23/25 (\phantom{0}92\%)} \\
Peg Insertion & 7/25 (28\%) & 12/25 (48\%) & \textbf{23/25 (\phantom{0}92\%)} \\
\midrule
\multicolumn{4}{@{}l}{\textit{Real world (R1 Pro, VLA)}} \\
Prepare Make-Ahead Breakfast Bowls & 1/25 (\phantom{0}4\%) & 19/25 (76\%) & \textbf{24/25 (\phantom{0}96\%)} \\
Boxing Books up for Storage & 3/25 (12\%) & 13/25 (52\%) & \textbf{23/25 (\phantom{0}92\%)} \\
Make Microwave Popcorn & 5/25 (20\%) & 20/25 (80\%) & \textbf{25/25 (100\%)} \\
Turning on Radio & 6/25 (24\%) & 14/25 (56\%) & \textbf{24/25 (\phantom{0}96\%)} \\
\bottomrule
\end{tabular}%
}
\end{table}

Table~\ref{tab:task_summary} compares three conditions: the base policy, training with 100 additional expert demonstrations, and training with 100 correction trajectories collected over four HG-DAgger rounds. Both augmented conditions start from the same base demonstration set. This comparison evaluates full-task reliability when the same additional trajectory budget is allocated to demonstrations or on-policy corrections.

Additional demonstrations improve base-policy success by 32.7 percentage points on average across the eight tasks, while four-round HG-DAgger improves it by 68.3 points, yielding a 35.6-point advantage over collecting 100 additional demonstrations. Despite starting from base policies with an average success rate of only 25.6\% across the eight tasks, I3L raises average success to 93.9\%, achieving at least 90\% on every task. In particular, on the four long-horizon real-world BEHAVIOR tasks~\citep{li2024behavior1k}, corrections improve success over the base policies by an average of 81 percentage points, outperforming additional demonstrations by 30 percentage points. These results show that, with the same training-trajectory budget, targeted corrections at deployment time reliably and more effectively close the last-mile gap in whole-body mobile manipulation than collecting additional demonstrations.

\subsection{Q2: Iteration Improves Performance and Reduces Intervention Burden}
\label{sec:q2}

\begin{figure*}[!t]
\centering
\includegraphics[width=1.0\textwidth]{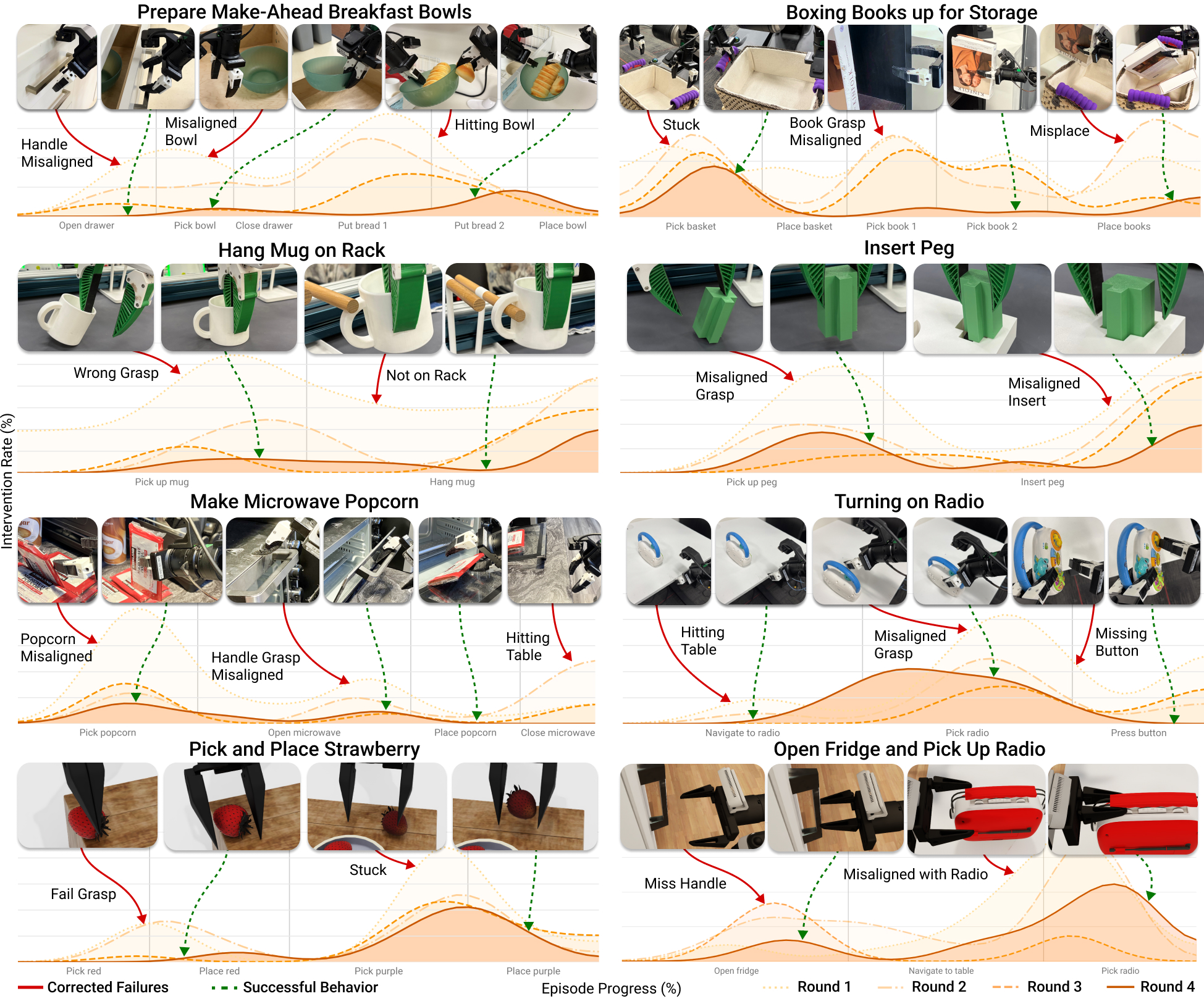}
\caption{\textbf{Task-stage intervention patterns across correction rounds.} Orange curves show intervention rate over normalized episode progress, where the color and line style distinguish Rounds~1--4. Each y-axis tick interval represents 10 percentage points. Red annotations indicate failures and green annotations indicate successful behaviors.}
\label{fig:corrections}
\end{figure*}

\begin{table}[!t]
\centering
\caption{Correction-learning comparison in simulation. Entries report mean $\pm$ sample standard deviation of success rates (\%) across five evaluation seeds. Parentheses indicate cumulative corrective trajectories. The entry with the best mean for each task is bold.}
\label{tab:algorithm_comparison}
\scriptsize
\setlength{\tabcolsep}{1pt}
\begin{tabular*}{\columnwidth}{@{\extracolsep{\fill}}lcccc@{}}
\toprule
Setting & Residual & BC (all data) & SIRIUS & HG-DAgger \\
\midrule
\multicolumn{5}{@{}l}{\textit{Open Fridge and Pick Up Radio (VLA)}} \\
Single round (100) & $21.6\pm17.5$ & $79.2\pm13.2$ & $73.2\pm12.5$ & $80.0\pm14.8$ \\
Round 1 (25) & $16.0\pm2.0$ & $70.8\pm4.6$ & $54.8\pm20.9$ & $51.6\pm12.3$ \\
Round 2 (50) & $24.0\pm5.1$ & $86.4\pm3.8$ & $68.4\pm13.1$ & $72.4\pm6.2$ \\
Round 3 (75) & $30.0\pm12.2$ & $87.2\pm6.6$ & $70.4\pm16.9$ & $82.0\pm13.6$ \\
Round 4 (100) & $19.6\pm6.2$ & $88.0\pm3.7$ & $71.6\pm7.9$ & $\boldsymbol{90.0\pm2.4}$ \\
\midrule
\multicolumn{5}{@{}l}{\textit{Pick-and-Place Strawberries (Diffusion Policy)}} \\
Single round (100) & $45.2\pm2.7$ & $14.8\pm2.3$ & $76.0\pm2.0$ & $89.2\pm3.0$ \\
Round 1 (25) & $49.2\pm4.1$ & $0.4\pm0.9$ & $70.8\pm8.9$ & $75.6\pm6.5$ \\
Round 2 (50) & $45.6\pm1.7$ & $1.2\pm1.1$ & $78.8\pm4.4$ & $88.4\pm3.8$ \\
Round 3 (75) & $40.0\pm7.6$ & $4.4\pm2.6$ & $37.6\pm6.8$ & $88.4\pm4.8$ \\
Round 4 (100) & $46.8\pm9.7$ & $10.8\pm5.0$ & $78.8\pm5.8$ & $\boldsymbol{92.8\pm2.7}$ \\
\bottomrule
\end{tabular*}
\end{table}

\paragraph{Iteration improves success under the same collection budgets}
Table~\ref{tab:algorithm_comparison} compares four iterative rounds of 25 episodes with a single batch of 100 episodes collected against the fixed base policy. With HG-DAgger filtering, iterative collection improves success by 3.6 percentage points on Pick-and-Place Strawberries and 10.0 points on Open Fridge and Pick Up Radio. Updating the policy between rounds changes the visited state distribution and allows later corrections to address failures of the updated policy.

\paragraph{Active intervention burden decreases across rounds}
Figure~\ref{fig:intervention_trends} shows a consistent decline in averaged intervention burden across the eight tasks. From Round~1 to Round~4, the mean number of takeovers per episode decreases by 63.4\%, the mean fraction of frames under intervention decreases by 69.5\%, and the mean total takeover duration per episode decreases by 72.0\%. Together, these measures indicate that takeovers become less frequent, occupy a smaller fraction of execution, and require less time under intervention per episode. All three task-averaged measures decline in every round, while the final policies achieve at least 90\% autonomous success on every task. 

\paragraph{Remaining bottlenecks are task dependent}
Although average intervention burden decreases, the remaining corrections concentrate around task-specific difficulties. Figure~\ref{fig:corrections} shows that on Boxing Books, intervention around book pickup decreases substantially by Round~4, while a prominent peak remains around basket pickup, possibly due to the difficulty of precisely coordinating both grippers with the basket handles with base, torso, and arm motion. On Prepare Breakfast, intervention during drawer opening and picking the first bread diminishes, while late-stage intervention remains around picking and placing the second smaller bread, which requires more precise gripper alignments. These two long-horizon BEHAVIOR tasks~\citep{li2024behavior1k} illustrate how iterative collection concentrates correction data on the difficult parts of each task where the policy continues to struggle.

\subsection{Q3: Corrections Span the Whole Body}
\label{sec:q3}

\begin{figure}[!t]
\centering
\setlength{\abovecaptionskip}{3pt}
\includegraphics[width=\columnwidth]{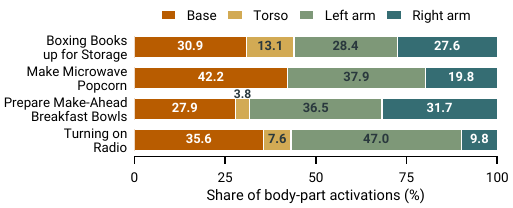}
\caption{\textbf{Body-part activation shares during JoyLo+ corrections.} Bars pool four 25-episode rounds; each body part receives one vote per takeover segment after three consecutive active frames. Thresholds are 0.003\,m/s for base translation and 0.003\,rad/s for base rotation commands, 0.01\,rad for torso joint differences, 0.02\,rad for arm joint differences, and 0.75 for normalized gripper differences. Grippers are assigned to their respective arms.}
\label{fig:wholebody}
\end{figure}

We analyze how operators use JoyLo+ during interventions on the four long-horizon real-world tasks. To characterize coordinated whole-body involvement, we identify sustained body-part activity within each takeover segment and normalize the resulting activation counts by task. Multiple body parts can contribute within the same segment.

Figure~\ref{fig:wholebody} shows that operators use JoyLo+ to correct more than arm motion: base activations account for an average of 34.2\% of body-part votes across all four tasks, while torso activations contribute an average of 8.2\% in the three tasks where they occur. Arm correction varies by task, with comparable left and right arm shares in Boxing Books and Prepare Breakfast, but larger left-arm shares in Turning on Radio and Make Popcorn. These patterns highlight the value of an interface that lets operators engage different combinations of the base, torso, and arms as the task requires.

\subsection{Q4: Correction-Learning Algorithm Comparison}
\label{sec:q4}

Table~\ref{tab:algorithm_comparison} compares HG-DAgger filtering with the following alternatives. Each method collects corrections with its own policy and trains on its accumulated data across rounds.

\subsubsection{Residual Policy} The residual policy uses a ResNet~\cite{kaimingresnet} image encoder and an MLP to predict additive action corrections from images, proprioception, and the base policy's predicted action chunk. The base policy remains frozen across rounds while the residual model is trained on accumulated corrections. The human demonstrations receive a zero residual target, and we sample correction timesteps with 50\% probability during training to balance the limited corrective data.

\subsubsection{BC (all data)} All collected data, including potentially suboptimal on-policy data, is used for training the policy.

\subsubsection{SIRIUS} Following the original SIRIUS sampling scheme~\citep{liu2023sirius}, intervention frames receive 50\% of the sampling probability, demonstrations receive up to 50\%, and any remaining probability is assigned to on-policy data.

Four-round HG-DAgger achieves the highest final mean success on both tasks. BC (all data) finishes close to HG-DAgger on Open Fridge, but performs poorly on Strawberries, consistent with learning suboptimal behavior from on-policy data. SIRIUS benefits less consistently from iterative collection: performance fluctuates across four rounds on Strawberries and it underperforms single-round collection on Open Fridge. Residual policy shows little final benefit from iteration relative to single-round collection on either task. This may reflect the difficulty of learning the large residuals required for some corrections. These results support multi-round HG-DAgger as a practical choice for iterative corrective learning across tasks and policies, while the benefits of additional rounds depend on how the collected data is used.

\subsection{Q5: Operator Study}
\label{sec:q5}

\begin{figure}[!htbp]
\centering
\setlength{\abovecaptionskip}{3pt}
\includegraphics[width=\columnwidth,trim={0 1.5pt 0 0},clip]{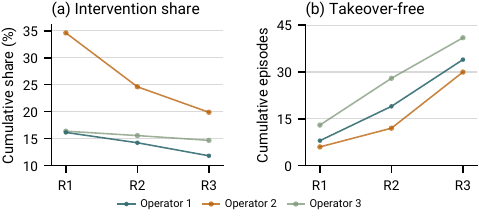}
\caption{\textbf{Operator study on Mug Hanging.} (a) Cumulative intervention share. (b) Cumulative takeover-free episode count. Both metrics aggregate episodes through the indicated round for each operator.}
\label{fig:operator-study}

\end{figure}

Prior work shows that differences in demonstrator behavior can affect imitation learning~\citep{gandhi2023compatible,mandlekar2022whatmatters}. Motivated by this variability, we examine whether I3L supports policy improvement and declining intervention burden across three teleoperators with varying levels of experience on Franka Mug Hanging. Each operator collects 25 episodes per round over three correction rounds, totaling 225 episodes. Figure~\ref{fig:operator-study} summarizes cumulative intervention burden and takeover-free episodes across their respective training histories. Cumulative intervention share decreases for all three operators, from 16.1\% to 11.8\%, 34.6\% to 19.9\%, and 16.4\% to 14.7\%. By Round~3, takeover-free episodes account for 40--55\% of each operator's cumulative collection. Together, these results suggest that I3L enables iterative policy refinements with declining intervention burden across users with varied teleoperation experience.

\section{Conclusion and Limitations}
We present \algo, an integrated hardware and learning system that combines JoyLo+ whole-body takeover with iterative interactive learning for long-horizon mobile manipulation. Across eight tasks, \algo consistently achieves at least 90\% autonomous success, outperforming demonstration-only training by an average of 35.6 percentage points under the same trajectory budgets. The mean takeover duration per episode decreases by 72\% across four rounds. Controlled simulation experiments show that the benefits of iterative collection depend on the learning strategy, with multi-round HG-DAgger achieving the highest final mean success.

Given the resource demands of multi-round whole-body experiments, our evaluation spans eight tasks, with controlled algorithm comparisons in simulation, real-world validation of the selected protocol, and a three-operator study. Broader task and operator coverage remains an important direction. Each round currently restarts training from a common initialization or pretrained checkpoint; incremental updates could reduce retraining cost while preserving performance. We measure active takeover burden, while operators continue to supervise execution. Future work will quantify total operator time and explore uncertainty- and failure-aware assistance.

\bibliographystyle{IEEEtran}
\bibliography{ref}  %

\clearpage
\useRomanappendicesfalse
\appendices
\setcounter{equation}{0}
\setcounter{table}{0}
\setcounter{figure}{0}
\renewcommand{\theequation}{A\arabic{equation}}
\renewcommand{\thetable}{A\arabic{table}}
\renewcommand{\thefigure}{A\arabic{figure}}
\newcommand{\appneed}[2]{\par\smallskip{\color{red!65!black}\noindent
  \textbf{Author input needed---#1.} #2\par}\smallskip}
\newcommand{\appcode}[1]{\texttt{\detokenize{#1}}}
\newcommand{\appreview}[1]{{\color{red!65!black}#1}}

\section{Robot Hardware, Observations, and Control}
\label{app:robot}

The following commands are expressed in robot units before the learning-side
normalization described in Appendix~\ref{app:policy}. Joint numbering is
one-based. Each topic distinguishes the two real robots from the two
simulation setups.

\subsection{Action Representations}
\label{app:actions}

\paragraph{Franka (Real)}
\label{app:robot-franka}
The eight-dimensional command is
\begin{equation}
  \mathbf a^{F}_t=(q^{\mathrm{cmd}}_{1:7,t},g^{\mathrm{cmd}}_t)
  \in\mathbb R^8.
  \label{app:eq:franka-action}
\end{equation}
The seven arm entries are absolute joint-position targets in radians,
followed by a gripper command. Negative gripper values request
opening and positive values request closing. The collection configuration
allows the full nominal 80\,mm opening at $-1$.
The diffusion policy consumes seven measured joint angles and the last
gripper command, giving eight state coordinates. The larger recorded
state additionally includes arm-velocity estimates
(Appendix~\ref{app:data}).

\paragraph{R1 Pro (Real)}
\label{app:robot-r1-real}
The 23-dimensional command is
\begin{equation}
 \mathbf a^{R}_t=(\mathbf b_{3,t},\mathbf q^{T}_{4,t},
 \mathbf q^{L}_{7,t},g^L_t,\mathbf q^{R}_{7,t},g^R_t).
 \label{app:eq:r1-action}
\end{equation}
The groups are base (3), torso (4), left arm (7), left gripper (1),
right arm (7), and right gripper (1). The base command is
$\mathbf b_{3,t}=(v_{x,t},v_{y,t},\omega_t)$, with two planar
velocities expressed in the robot's base frame in m/s and yaw
velocity in rad/s. Torso and arm
entries are absolute joint-position targets in radians; grippers use
$[-1,1]$, with $-1$ open and $+1$ closed.
The ROS~2 \appcode{brs-ctrl} adapter clips joint targets and base velocities.
Table~\ref{app:tab:r1-limits} gives its base and torso limits.
Base commands below 0.01\,m/s on either translation axis or 0.05\,rad/s in
yaw are set to zero.

\begin{table}[!htbp]
\centering
\caption{Real-R1 adapter limits. Base bounds apply independently
to each component; torso rows follow the robot's joint ordering.}
\label{app:tab:r1-limits}
\footnotesize
\setlength{\tabcolsep}{6pt}
\begin{tabular}{@{}llrr@{}}
\toprule
Command & Unit & Lower & Upper \\
\midrule
Base translation $x$ & m/s & $-0.3$ & $0.3$ \\
Base translation $y$ & m/s & $-0.3$ & $0.3$ \\
Base yaw & rad/s & $-0.4$ & $0.4$ \\
Torso joint 1 & rad & $-1.1345$ & $1.8326$ \\
Torso joint 2 & rad & $-2.7925$ & $2.5307$ \\
Torso joint 3 & rad & $-1.8326$ & $1.5708$ \\
Torso joint 4 & rad & $-3.0543$ & $3.0543$ \\
\bottomrule
\end{tabular}
\end{table}

The real-R1 VLA configuration selects 26 state coordinates:
base pose (3), torso angles (4), both arm configurations (14), both gripper
openings mapped to the command range (2), and base velocity (3).
This selection is made from the 46 recorded state/velocity channels
(Appendix~\ref{app:data}) and padded to 32 for the model.

\paragraph{A1 (Simulation)}
\label{app:robot-a1-sim}
The simulated A1 uses the seven-dimensional command
\begin{equation}
  \mathbf a^{A}_t=(q^{\mathrm{cmd}}_{1:6,t},g^{\mathrm{cmd}}_t)
  \in\mathbb R^7.
  \label{app:eq:a1-action}
\end{equation}
The six arm entries are absolute joint-position targets in radians;
the gripper command is in $[-1,1]$, with $+1$ open and $-1$ closed.
Arm joints use position control without an additional impedance controller;
the gripper uses smooth position control and physical grasping.
The diffusion policy uses six measured arm angles and a binary gripper
state derived from the mean finger position: above 0.024\,m is open in
the current preprocessing.

\paragraph{R1 Pro (Simulation)}
\label{app:robot-r1-sim}
The simulated R1~Pro uses the 23-dimensional command layout in
Eq.~\eqref{app:eq:r1-action}, with
$\mathbf b_{3,t}=(\Delta x_t,\Delta y_t,\Delta\psi_t)$ specifying
local-frame translation and yaw increments in metres and radians per
action step. The controller
converts these increments to position targets. Torso and arm commands are
absolute joint angles in radians; grippers use $+1$ open and $-1$ closed.
Base, torso, and arm controllers use position control without an additional
impedance controller. Grippers use smooth position control with assisted
grasping.

The simulation VLA configuration selects 23 state coordinates:
base velocity (3), torso angles (4), both arms (14), and the summed finger
width for each gripper (2), then pads to 32. This selection excludes base
pose.

\FloatBarrier
\subsection{Home Poses and Kinematic Calibration}
\label{app:calibration}

\paragraph{Franka (Real)}
The nominal arm home pose and calibrated leader offsets are given in
Table~\ref{app:tab:franka-calibration}; the home command requests an open gripper.
Encoder angles use a zero of 2048 counts and a scale of $\pi/2048$
radians per count, followed by the listed sign and additive offset.
For peg insertion, joint~7 instead homes at $-3\pi/4$, with an additional
$-\pi/2$ leader offset so that the leader retains its physical home posture.
These calibration values are specific to the leader configuration.

\begin{table}[!htbp]
\centering
\caption{Franka mapping and nominal home pose. Offsets and home
angles are in radians; the insertion adjustment is applied separately.}
\label{app:tab:franka-calibration}
\footnotesize
\setlength{\tabcolsep}{7pt}
\begin{tabular}{@{}crrr@{}}
\toprule
Joint & Sign & Leader offset & Home angle \\
\midrule
1 & $+1$ & $-1.3962$ & $0$ \\
2 & $+1$ & $ 1.6046$ & $0$ \\
3 & $+1$ & $ 3.1416$ & $0$ \\
4 & $-1$ & $-2.3000$ & $-\pi/2$ \\
5 & $+1$ & $-3.2681$ & $0$ \\
6 & $-1$ & $ 1.6402$ & $\pi/2$ \\
7 & $+1$ & $-0.7000$ & $-\pi/4$ \\
\bottomrule
\end{tabular}
\end{table}

\paragraph{R1 Pro (Real)}
The four torso joints are initialized to $(0,0,0,0)$ radians.
The final base home command is zero translational
velocity along both horizontal axes (0\,m/s) and zero yaw rate (0\,rad/s);
it does not specify a fixed base position or heading. In the real tasks,
the base is initialized within the task-specific yellow-mat regions shown
in Fig.~\ref{fig:tasks}.
The leader map has zero additive offsets, with joint signs
$(-1,-1,-1,-1,-1,-1,+1)$ on the right and
$(+1,-1,-1,+1,-1,-1,+1)$ on the left; encoder zero and scale are as for
Franka.

\paragraph{A1 (Simulation)}
The strawberry task sets the arm to
$(0,1.877,-0.945,1.570,0.931,0)$ radians and each finger coordinate to
0.03\,m. The robot remains at a fixed base pose while objects are randomized
as described in Appendix~\ref{app:simulation}.

\paragraph{R1 Pro (Simulation)}
For Fridge/Radio, the task sets the torso to $(-0.63,1.65,1.05,0)$ radians
and each arm to $(0,0,0,-1.57,0,0,0)$ radians, with each finger coordinate
at 0.05\,m. The initial base position varies by $\pm0.15$\,m along each
horizontal axis and heading by $\pm18^\circ$.
Scene and object randomization are specified in Appendix~\ref{app:simulation}.

\FloatBarrier
\subsection{Camera Layout and Image Formation}
\label{app:cameras}

\paragraph{Franka (Real)}
An external ZED~2i and wrist-mounted ZED Mini each capture side-by-side
stereo at $2560\times720$ pixels and 30\,Hz. Only the left eye is selected,
yielding $1280\times720$ pixels. Table~\ref{app:tab:franka-cameras} specifies
the square crops; offsets are relative to the image centre, positive right
and down. Each crop is resized to $240\times240$ RGB. The policy then takes
a random $224\times224$ crop during training and a centred crop during
evaluation, with no depth input.

\begin{table}[!htbp]
\centering
\caption{Recorded Franka camera profile. Crop widths and offsets are in pixels,
before resizing to $240\times240$.}
\label{app:tab:franka-cameras}
\footnotesize
\setlength{\tabcolsep}{4pt}
\begin{tabular}{@{}llrr@{}}
\toprule
View & Camera & Crop width & Offset $(x,y)$ \\
\midrule
External & ZED 2i & 600 & $(-30,-60)$ \\
Wrist & ZED Mini & 720 & $(0,0)$ \\
\bottomrule
\end{tabular}
\end{table}

Both cameras use ZED SDK~5.1.0 and firmware~1523, with the same camera setup
across all reported rounds. The appearance profile uses automatic wrist
exposure/gain and external exposure~22 with gain~0 in SDK units. Saved
reference images support visual alignment of the views.

\paragraph{R1 Pro (Real)}
The cameras installed on the Galaxea R1~Pro provide head, left-wrist,
and right-wrist RGB views. The
controller's default head input selects the left image; its wrist images
are rectified. The local interface specifies $240\times240$ images, while
the training datasets store $224\times224$ views. The VLA applies
aspect-ratio-preserving resize and padding to $224\times224$ during both
training and inference and also receives a task-language prompt.

\paragraph{A1 (Simulation)}
External and wrist RGB images are rendered at $240\times240$ and cropped
to $224\times224$ for the policy, randomly during training and centrally
during evaluation. Table~\ref{app:tab:a1-cameras} specifies the configured
camera poses; no physical camera SDK or firmware is involved.

\begin{table}[!htbp]
\centering
\caption{A1 simulation camera poses. Translation is in metres and
quaternions use $(x,y,z,w)$ order. The external pose is relative to its scene
parent and the wrist pose to the end-effector link.}
\label{app:tab:a1-cameras}
\scriptsize
\setlength{\tabcolsep}{3pt}
\begin{tabular}{@{}lll@{}}
\toprule
View & Translation & Quaternion \\
\midrule
External & $(0.7,0,0.6)$ & $(0.255,0.255,0.66,0.66)$ \\
Wrist & $(0.086,0,-0.066)$ & $(0.652,0.652,0.274,0.274)$ \\
\bottomrule
\end{tabular}
\end{table}

\paragraph{R1 Pro (Simulation)}
Head and both wrist RGB views are rendered at $240\times240$ and processed
to $224\times224$ for the VLA, together with the task prompt.
The camera mounts follow the simulated robot asset.

\FloatBarrier
\noindent\begin{minipage}{\columnwidth}
\subsection{Frequencies, Timing, and Input Devices}
\label{app:timing}

Table~\ref{app:tab:rates} summarizes the configured process rates.
Environment stepping, camera capture, simulation physics, leader control,
and touch-state publication run on separate clocks. Leader-control and
touch-sensing timing are detailed in Appendix~\ref{app:joylo}.

\begin{table}[H]
\centering
\caption{Configured operating rates for the real and simulated setups.}
\label{app:tab:rates}
\footnotesize
\begin{tabular}{@{}lr@{}}
\toprule
Process & Configured rate \\
\midrule
Real Franka/R1 Pro environment loop & 30\,Hz \\
Franka camera capture & 30\,Hz \\
Simulated action / rendering & 30 / 30\,Hz \\
Simulated physics & 120\,Hz \\
JoyLo+ leader control & 100\,Hz \\
Touch-state serial publication & 200\,Hz \\
\bottomrule
\end{tabular}
\end{table}
\end{minipage}\par\medskip

\paragraph{Franka (Real)}
The robot is controlled through Deoxys~\citep{zhu2022viola}. The
wrapper configures its client at 100\,Hz, with proportional/derivative
settings of 200/14 for each arm joint across all reported rounds;
the environment action loop runs at 30\,Hz. Two consecutive observations
condition each prediction; eight actions are executed before replanning
(Appendix~\ref{app:policy}). The leader supplies arm commands, and the
right Joy-Con controls the gripper and trial recording
(Table~\ref{app:tab:franka-joycon}).

\begin{table}[!htbp]
\centering
\caption{Franka Joy-Con controls for teleoperation and trial recording.}
\label{app:tab:franka-joycon}
\footnotesize
\setlength{\tabcolsep}{3pt}
\renewcommand{\arraystretch}{1.1}
\begin{tabular}{@{}p{0.25\columnwidth}p{0.71\columnwidth}@{}}
\toprule
Control & Function \\
\midrule
R & Close gripper \\
ZR & Open gripper \\
X & Start recording \\
A & Accept and save trial \\
Y & Discard trial \\
Home & Home robot and leader \\
$+$ & End collection block \\
\bottomrule
\end{tabular}
\end{table}

\paragraph{R1 Pro (Real)}
The environment loop targets 30\,Hz; the adapter configures joint
interpolation at 100\,Hz. The local wrapper uses 3.0\,rad for the arms
and 1.0\,rad for the torso when determining the number of interpolation
steps.

Leaders control the arms, while the Joy-Cons control the base, torso,
grippers, and trial recording (Table~\ref{app:tab:r1-joycon}).
The input configuration uses a 20\% joystick dead zone and exponential
smoothing coefficient 0.9.

\begin{table}[!htbp]
\centering
\caption{R1 Pro Joy-Con controls for teleoperation and trial recording.
Torso angles are in radians.}
\label{app:tab:r1-joycon}
\footnotesize
\setlength{\tabcolsep}{3pt}
\renewcommand{\arraystretch}{1.1}
\begin{tabular}{@{}p{0.35\columnwidth}p{0.61\columnwidth}@{}}
\toprule
Control & Function \\
\midrule
Left stick, unpressed & Base translation (up to 0.3\,m/s per axis) \\
Left stick, pressed & Base yaw via horizontal deflection (up to 0.4\,rad/s) \\
Right stick, vertical & Torso pitch (joint 3) \\
Right stick, horizontal & Torso yaw (joint 4) \\
Left D-pad, up / down & Move torso joints 1--2 toward stand $(0,0)$ / squat $(1.74,-2.70)$ \\
ZL / ZR & Toggle left / right gripper \\
X / A / Y & Start recording / save / discard trial \\
Home & Home robot and leaders \\
$+$ & End collection block \\
\bottomrule
\end{tabular}
\end{table}

\paragraph{A1 and R1 Pro (Simulation)}
Both simulations apply actions and render RGB images at 30\,Hz, with
physics at 120\,Hz. Keyboard input selects the scripted correction oracle;
the platform-specific intervention protocols are described in
Appendix~\ref{app:takeover}.

\FloatBarrier
\section{JoyLo+ Mechanical and Takeover Design}
\label{app:joylo}

JoyLo+ extends GELLO~\citep{wu2023gello} and JoyLo~\citep{jiang2025brs}
with actuated policy tracking and capacitive-touch takeover for the real
robots. The simulated tasks use human-triggered scripted corrections.
Each topic below distinguishes the real and simulated interfaces.

\subsection{Mechanical Structure and Actuation}

Both real-robot interfaces use the same leader-control architecture.
A 100\,Hz loop sends policy joint targets to the motors' onboard position
controllers during autonomous operation. During human takeover, the motors
switch to current control (mode~0) with model-based gravity compensation
for hand guidance. The software bounds takeover current commands by each
motor's stored current limit. The configurations below specify the
autonomous tracking-current caps and hand-guidance tuning for each robot.

\paragraph{Franka (Real)}
\label{app:joylo-franka}
The seven-joint leader uses three XM430-W350-T servos for robot joints
1, 2, and 4, and four XC330-M288-T servos for joints 3, 5, 6, and 7
(one-based numbering). Each motor family has a separate USB serial bus
at 1\,Mbit/s, with the larger motors on a 12\,V supply and the smaller
motors on a 5\,V supply. The printed linkage, motor-to-joint assignment,
and calibrated offsets reproduce the Franka arm's joint layout.

For autonomous position tracking, the Franka configuration does not add
a tracking-current cap. For hand guidance, the software proportional,
integral, and damping gains are zero. The empirically determined,
dimensionless gravity multipliers, in joint order 1--7, are
$(0,-0.4,0.5,-0.5,0.3,-0.3,0)$.
Velocity-proportional assistance
supplements gravity compensation, with per-joint coefficients
$(0.01,0.02,0.01,0.02,0.03,0.01,0.01)$\,N\,m\,s/rad and a
0.005\,rad/s velocity deadband.

\paragraph{R1 Pro (Real)}
\label{app:joylo-r1-real}
R1~Pro uses two mirrored seven-joint leaders. On each leader,
joints 1--4 use XM430-W350-T motors and joints 5--7 use XC330-M288-T
motors. Both leaders share two serial buses total: one connects all
eight larger motors and the other all six smaller motors, each at
1\,Mbit/s. The larger-motor bus uses 12\,V power and the smaller-motor
bus 5\,V power, with a USB adapter for each bus.

For autonomous position tracking, the R1~Pro configuration adds
per-motor tracking-current caps:
400 raw current units (1.076\,A) per larger motor and 150 units
(0.150\,A) per smaller motor. For hand guidance, the software uses
reduced proportional and damping gains with zero integral gain.
The gravity-compensation model accounts for the mirrored mounting.
Its empirically determined, dimensionless gravity multipliers, in joint
order 1--7, are
$(0.4,0.4,0.1,0.1,0,0.3,0.01)$ for the right arm and
$(0.3,0.3,0.1,0.1,0,0.3,0)$ for the left arm.

\paragraph{A1 and R1 Pro (Simulation)}
These collection interfaces use scripted corrections and do not require
physical JoyLo+ leaders or motor buses. Their correction commands are
described in Appendix~\ref{app:takeover}.

\subsection{Touch Hardware, Wiring, and Firmware}
\label{app:touch}

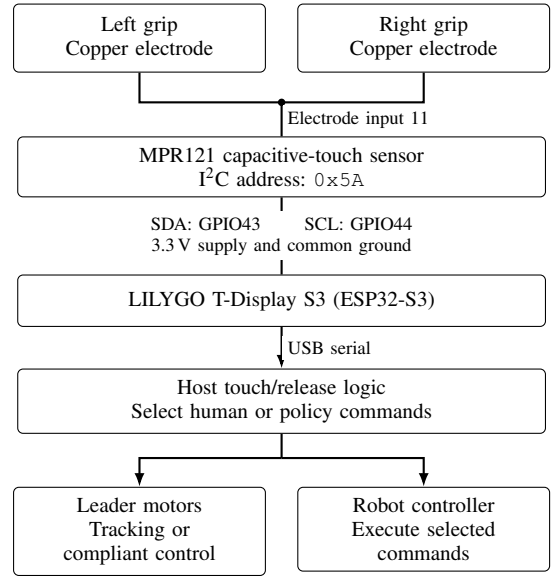
\begin{figure}[t]
\centering
\begin{tikzpicture}[
  font=\footnotesize,
  block/.style={draw, rounded corners=2pt, align=center,
    text width=6.7cm, minimum height=0.68cm, inner sep=4pt},
  endpoint/.style={block, text width=3.05cm, minimum height=0.9cm},
  flow/.style={-{Latex[length=1.8mm]}, thick},
  wire/.style={thick},
  note/.style={font=\scriptsize, align=center, fill=white, inner sep=2pt}
]
\node[endpoint] (leftgrip) at (-1.88cm,0)
  {Left grip\\Copper electrode};
\node[endpoint] (rightgrip) at (1.88cm,0)
  {Right grip\\Copper electrode};
\coordinate (gripjoin) at (0,-0.85cm);
\draw[wire] (leftgrip.south) |- (gripjoin);
\draw[wire] (rightgrip.south) |- (gripjoin);
\fill (gripjoin) circle (1.4pt);
\node[block, anchor=north] (sensor) at (0,-1.30cm)
  {MPR121 capacitive-touch sensor\\I\textsuperscript{2}C address: \texttt{0x5A}};
\draw[wire] (gripjoin) -- node[note, right] {Electrode input 11} (sensor.north);

\node[block, below=1.02cm of sensor] (mcu)
  {LILYGO T-Display S3 (ESP32-S3)};
\draw[wire] (sensor.south) -- node[note]
  {SDA: GPIO43 \qquad SCL: GPIO44\\3.3\,V supply and common ground} (mcu.north);

\node[block, below=0.55cm of mcu] (host)
  {Host touch/release logic\\Select human or policy commands};
\draw[flow] (mcu.south) -- node[note, right] {USB serial} (host.north);

\node[endpoint, below=0.7cm of host, xshift=-1.88cm] (leader)
  {Leader motors\\Tracking or compliant control};
\node[endpoint, below=0.7cm of host, xshift=1.88cm] (robot)
  {Robot controller\\Execute selected commands};
\coordinate (split) at ($(host.south)+(0,-0.32cm)$);
\draw[wire] (host.south) -- (split);
\draw[flow] (split) -| (leader.north);
\draw[flow] (split) -| (robot.north);
\end{tikzpicture}
\caption{R1 Pro touch wiring and command flow. Both grip electrodes connect
to input 11 of one MPR121 and share one ESP32-S3; touching either grip
activates takeover.}
\label{app:fig:touch-wiring}
\end{figure}

\paragraph{Shared electronics: Franka and R1 Pro (Real)}
Each real interface uses one MPR121 touch-sensor board and one
LILYGO T-Display S3 (ESP32-S3). Copper tape on the grip or grips connects
to electrode input~11.
The ESP32-S3 reads the MPR121 at I$^2$C address
\appcode{0x5A}: SDA connects to GPIO~43, SCL to GPIO~44, and the sensor
uses 3.3\,V and common ground. Figure~\ref{app:fig:touch-wiring} shows
the R1~Pro wiring. Touch on any enabled electrode is treated as contact.

Table~\ref{app:tab:touch} gives the touch-sensor settings. The host
receives touch states over USB at a configured 200\,Hz
(115200-baud serial setting); the sensor's electrode sampling interval
is 16\,ms. Sensor sampling, debounce, publication, and release delay are distinct
stages.

\begin{table}[!htbp]
\centering
\caption{MPR121 touch-sensor settings. Thresholds are in sensor register units.}
\label{app:tab:touch}
\footnotesize
\setlength{\tabcolsep}{3pt}
\renewcommand{\arraystretch}{1.1}
\begin{tabular}{@{}p{0.31\columnwidth}p{0.65\columnwidth}@{}}
\toprule
Setting & Value \\
\midrule
Enabled electrodes & All 12 channels \\
Thresholds\newline(touch / release) & 195 / 117 \\
Charge current & 63\,$\mu$A \\
Sample interval & 16\,ms \\
Debounce\newline(touch / release) & Three consecutive
qualifying detections \\
Baseline calibration & Automatic configuration and tracking \\
\bottomrule
\end{tabular}
\end{table}

The calibration procedure records 1,000 samples with the
electrode untouched and 1,000 while firmly held, discarding 50 settling
readings before each set. From baseline-minus-filtered sensor readings,
it places the touch threshold 30\% of the way from the untouched
99th percentile to the touched 5th percentile; the release threshold
is 60\% of the touch threshold. The resulting integer thresholds are
written to firmware and require reflashing. A disconnected sensor or
touch data older than 0.5\,s is treated as inactive and follows the
normal release path if no other input is active.

\paragraph{Franka (Real)}
The single grip connects to electrode input~11. The Franka host retains
human authority for 0.5\,s after touch ceases,
unless touch or another active operator input renews it.

\paragraph{R1 Pro (Real)}
The left and right grip electrodes are wired together to input~11 on the
shared MPR121 board, read by a single ESP32-S3. Touching either grip
activates the same takeover signal. The host retains human authority for
0.3\,s after both grips are released, unless another active operator input
renews it.

\paragraph{A1 and R1 Pro (Simulation)}
The simulation interfaces use keyboard input, with no capacitive grips,
touch thresholds, or grip-release delays.

\subsection{Authority Switching and Policy Resumption}
\label{app:takeover}

\paragraph{Franka (Real)}
A touch or active JoyCon command selects the human arm/gripper command
immediately. The leader switches to compliant current control while the
robot follows the mapped leader pose and operator gripper command.
Human control continues throughout the release delay and the release
step. Release requests a new policy plan from the corrected observations;
the Franka implementation discards pending plans at both
takeover and release. At release, each leader's target is first set to
its measured pose before position tracking is restored. While waiting
for a fresh prediction, the robot holds rather than executing an obsolete
plan; leaders follow the new autonomous targets as policy execution resumes.
Policy predictions
remain enabled during correction, but temporary hold commands are not
valid counterfactual predictions.

\paragraph{R1 Pro (Real)}
Either grip contact or active JoyCon input transfers authority over
both arms, both grippers, the torso, and the base together. The leaders
supply the arm targets; JoyCons provide the remaining commands as
described in Appendix~\ref{app:robot}. The operator may move one
body part while holding the others under the same authority.

After release, human commands are retained for the release step and
the policy replans from the corrected state before subsequent autonomous
execution. At takeover entry, commands transition linearly from the policy
command to the human command over 2\,s.

\paragraph{A1 (Simulation)}
\label{app:joylo-a1-sim}
For Strawberries, the operator uses keyboard input to interrupt the
learned policy and activate a scripted oracle. The oracle uses
simulator task state and cuRobo planning~\citep{sundaralingam2023curobo}
to generate the six arm-joint targets and gripper command. The human
chooses when to intervene and when to resume policy execution; the
corrective motion itself is generated by the oracle. The learned policy
receives camera and proprioceptive observations and replans from the
corrected state. The task's oracle sequence is listed
in Appendix~\ref{app:collection}.

\paragraph{R1 Pro (Simulation)}
\label{app:joylo-r1-sim}
For Fridge and Radio, the same keyboard-triggered protocol activates a
scripted oracle for the complete simulated R1~Pro command, including
base, torso, arms, and grippers. The oracle chooses its next motion
from the current task state, allowing correction of door opening,
navigation, and radio grasping before returning to the learned policy.
Intervention labels and executed
commands are recorded throughout; Appendix~\ref{app:collection}
specifies the motion sequence and Appendix~\ref{app:labels} the labeling
convention.

\subsection{Bill of Materials and Cost Scope}
\label{app:bom}

For CAD models, refer to the project website at
\texttt{https://i-3l.github.io/}. Table~\ref{app:tab:bom} gives the estimated
component costs for both leader configurations.

\paragraph{Franka (Real)}
The single-arm interface uses three XM430-W350-T and four XC330-M288-T
servos. The estimated cost of the leader and listed accessories is
\$1,087.49.

\paragraph{R1 Pro (Real)}
Each leader uses four XM430-W350-T and three XC330-M288-T servos,
with an estimated cost of \$1,257.49 including the listed accessories.
The dual-leader interface uses eight XM430 and six XC330 servos and costs
approximately twice the single-leader estimate, or \$2,515.

\begin{table}[H]
\centering
\caption{Estimated bills of materials for one Franka or R1~Pro leader
and the listed accessories. Costs are in USD. Robot, tax, shipping,
cabling, fasteners, bearings, and mounting hardware are excluded.}
\label{app:tab:bom}
\footnotesize
\setlength{\tabcolsep}{3pt}
\renewcommand{\arraystretch}{1.1}
\begin{tabular*}{\columnwidth}{@{\extracolsep{\fill}}lrrr@{}}
\toprule
& & \multicolumn{2}{c}{Quantity} \\
\cmidrule(l){3-4}
Component & Unit cost & Franka & R1~Pro \\
\midrule
XM430-W350-T servo & \$220 & 3 & 4 \\
XC330-M288-T servo & \$50 & 4 & 3 \\
U2D2 USB adapter & \$30 & 2 & 2 \\
Power supply (12\,V/5\,V) & \$25 & 2 & 2 \\
Printed links (set) & \$15 & 1 & 1 \\
Joy-Con pair & \$80 & 1 & 1 \\
Adafruit MPR121 & \$7.95 & 1 & 1 \\
LILYGO T-Display S3 & \$9.54 & 1 & 1 \\
Copper tape & \$5 & 1 & 1 \\
\midrule
\multicolumn{2}{@{}l}{Estimated total} & \$1,087.49 & \$1,257.49 \\
\bottomrule
\end{tabular*}
\end{table}

\paragraph{A1 and R1 Pro (Simulation)}
The physical leader component costs above do not apply to these
keyboard-triggered simulation interfaces.

\FloatBarrier
\section{Simulation Setup and Data Collection}
\label{app:simulation}

\subsection{Simulator Configuration}

We run both simulation tasks in OmniGibson using BEHAVIOR
assets~\citep{li2024behavior1k,ge2024behaviorvisionsuite}. The A1 environment uses an empty scene,
while the R1~Pro
environment uses scene \appcode{Rs_int}, with selected scene objects and
the assets needed for the task.
Physics runs at 120\,Hz, while images are rendered and robot actions are
applied at 30\,Hz. The arm controllers receive absolute joint-position
targets. A1 uses external and wrist RGB cameras at $240\times240$
resolution; R1~Pro uses its three robot-mounted cameras. A1 uses physical
grasping, while R1~Pro uses assisted grasping. Both tasks use a 20-step
goal hold threshold, with timeouts of 800 action steps for Strawberries
and 900 for Fridge/Radio.

\subsection{Task Initialization}

\paragraph{Strawberries}
Strawberry positions are randomized at the start of each episode, while
the bowl and the robot's initial pose remain fixed. Initial
placements are sampled to maintain at least 0.15\,m separation from each
other and the bowl.

\paragraph{Fridge and radio}
Each episode starts with the refrigerator closed and the radio on a
separate table. The robot's initial base position varies by up to
$\pm0.15$\,m along each horizontal axis, and its heading varies by
$\pm18^\circ$. The refrigerator and table share a random horizontal
offset of up to $\pm0.10$\,m along one axis, with independent offsets of
up to $\pm0.05$\,m along the other. 

\subsection{Demonstration and Correction Collection}
\label{app:collection}

Scripted oracles use simulator task state and cuRobo motion
planning~\citep{sundaralingam2023curobo} to generate initial demonstrations
and corrective trajectories. During correction collection, a human operator
uses keyboard input to interrupt the learned policy, activate oracle-generated
corrections, and resume policy execution when the correction is complete. The learned policy
receives camera and proprioceptive observations and replans from the
corrected state after each intervention. Executed actions and intervention
labels are recorded throughout each rollout.
The oracle selects the next motion from the current task state.
These motion steps are grouped into the annotated subtasks shown in
Figure~\ref{fig:corrections}, where each subtask can contain several steps.
The nominal sequences are:

\paragraph{Strawberries (14 motion substeps)}
Open the gripper, move above the red strawberry,
lower to the grasp position, close the gripper, lift the red strawberry, move above the
bowl, and release, then open the gripper, move above the purple strawberry,
lower to the grasp position, close the gripper,
lift the purple strawberry, move above the bowl, and release.

\paragraph{Fridge and radio (14 motion substeps)}
Open the gripper, position the base and arm near the refrigerator handle,
reach the handle, grasp it, pull the door open, and release the handle.
Reset the torso, retract the arms, open the gripper, and navigate to the
radio table. Move above the radio handle, reach the handle,
grasp it, and lift the radio.

\FloatBarrier
\section{Data Recording and Processing}
\label{app:data}

\subsection{Recorded Data and Intervention Labels}
\label{app:labels}

Each rollout records camera RGB observations, robot proprioception, executed
commands, policy-proposed actions, and intervention labels. The executed
command follows the operator or scripted oracle during takeover and the
policy during autonomous execution. A correction trajectory therefore
contains both autonomous and intervention segments.

\paragraph{Franka}
Recordings contain two $240\times240$ RGB views from the external and wrist
cameras, and an 8-dimensional action vector comprising seven arm-joint
position targets and one gripper command. The 16-dimensional proprioceptive
vector contains seven measured joint angles, the last gripper command,
seven joint velocities estimated from consecutive measurements, and a zero
placeholder for gripper velocity.

\paragraph{Real R1~Pro}
The recording interface uses three $240\times240$ RGB views from the head
and two wrist cameras. Its 23-dimensional action vector consists of three
base velocity commands, four torso-joint targets, seven joint targets per arm, and
two gripper commands. The 46-dimensional proprioceptive vector concatenates
23 state entries and 23 velocity entries in the same body-part order.
Base and joint states and velocities come from robot feedback;
gripper opening measurements are mapped to the policy's gripper-command range.
The two gripper-velocity entries are zero placeholders.

\paragraph{Intervention labels}
We distinguish original demonstrations ($s_t=1$), active interventions
($s_t=2$), pre-intervention frames ($s_t=3$), and autonomous execution
($s_t=4$). The pre-intervention window includes the 15 frames immediately
before a takeover, and active intervention
labels take precedence when windows overlap. For takeover indicator $h_t$,
the labels are assigned as
\begin{equation}
 s_t=\begin{cases}
 1,&\text{original demonstration},\\
 2,&h_t=1,\\
 3,&h_t=0,\ \exists j\in\{t+1,\ldots,t+15\}:h_j=1,\\
 4,&\text{otherwise}.
 \end{cases}
 \label{app:eq:label-rule}
\end{equation}
These labels identify the source of supervision for training-data selection.

\subsection{Training-Data Preparation}
\label{app:processing}

\paragraph{Normalization}
For diffusion policies, joint values are scaled to $[-1,1]$ using the
configured robot-specific limits. For a value $q$ with range
$[q_{\min},q_{\max}]$, the transformation is
\begin{equation}
 \widetilde q=2\frac{q-q_{\min}}{q_{\max}-q_{\min}}-1.
 \label{app:eq:normalization}
\end{equation}
Non-gripper action groups use the same transformation with their configured
ranges; gripper channels use their platform-specific command encoding.
VLA data adaptation is described in Appendix~\ref{app:policy}.

\paragraph{Sampling by supervision source}
Let $N_c$ count eligible training-sample starts in supervision category $c$.

\textbf{HG-DAgger.} The sampler~\citep{kelly2019hgdagger} selects starts
from original demonstrations and active interventions. With $N_D$ eligible
demonstration starts and $N_I$ eligible intervention starts, their sampling
probabilities are
\begin{equation}
 p_D=\frac{N_D}{N_D+N_I},\qquad
 p_I=\frac{N_I}{N_D+N_I},\qquad p_P=p_A=0,
 \label{app:eq:hg-sampling}
\end{equation}
where $P$ and $A$ denote pre-intervention and autonomous frames.
The correction sampling fraction therefore depends on the available data.

\textbf{BC (all data).} We sample uniformly from all eligible chunk starts,
preserving the observed proportions of the four supervision categories:
\begin{equation}
 p_c=\frac{N_c}{N_D+N_I+N_P+N_A},\qquad c\in\{D,I,P,A\},
 \label{app:eq:bc-sampling}
\end{equation}

\textbf{SIRIUS.} Our SIRIUS sampler~\citep{liu2023sirius} uses the following
probabilities when demonstration, intervention, and autonomous data are
all available:
\begin{equation}
\begin{aligned}
 p_D&=\min(f_D,0.5), & p_I&=0.5,\\
 p_P&=0, & p_A&=0.5-p_D,
\end{aligned}
 \label{app:eq:sirius-sampling}
\end{equation}
where $f_D$ is the demonstration fraction among eligible chunk starts
before reweighting. Our implementation caps this fraction at 0.5.
If a category is absent, its weight is zero; any remaining weight is
assigned to autonomous data when available, and the weights are normalized
over the available categories.

\subsection{Subtask Annotation}
\label{app:subtasks}

Human annotators review the recorded episode videos using an interactive
annotation tool and mark each subtask's start and end frames. The tool saves
the named intervals in JSON files associated with each episode. Subtask names
follow the sequences shown in Figure~\ref{fig:corrections}. Note that these annotations
are used for analysis, while the separate intervention labels are used to
select data for training.

\paragraph{Intervention-rate curves}
For each task and collection round, we normalize frame positions by episode
length and divide whole-episode progress into 50 bins. Within each bin, we
pool intervention and total frame counts across episodes and compute their
ratio as a percentage. Takeover segments shorter than 15 frames are excluded
from the intervention counts, while their frames remain in the denominator.
The resulting rates are smoothed with a Gaussian filter with standard
deviation three bins and reflected boundary handling.

\paragraph{Displayed subtask boundaries}
The human annotations determine the subtask boundaries shown alongside the
curves, while recorded intervention labels determine the intervention rates.
Each displayed boundary is the mean subtask end-frame position, normalized
by episode length, across episodes in a reference round. Figure~\ref{fig:corrections}
uses Round~2 boundaries for Boxing Books up for Storage and Make Microwave
Popcorn, and Round~4 boundaries for the remaining tasks. Thus, all curves
use normalized whole-episode progress as their horizontal axis.

\FloatBarrier
\section{Policy Learning and Deployment}
\label{app:policy}

A1 and Franka use diffusion policies (DP)~\citep{chi2023diffusion} trained
on task data only; R1~Pro uses the pretrained $\pi_{0.5}$
VLA~\citep{intelligence2025pi05}. Every round retrains on the accumulated
data, restarting from the same initialization
(ImageNet encoder weights for DP, the $\pi_{0.5}$ base checkpoint for the
VLA). The final
checkpoint is evaluated after 200k training steps for DP and 40k steps for
the VLA under the protocol of Section~\ref{sec:setup}. Across experiments,
training used NVIDIA H100, H200, RTX PRO 6000, RTX 4090, RTX 3090, and
TITAN RTX GPUs. All real-robot policies are deployed on machines equipped
with NVIDIA RTX 4090 GPUs.
Tables~\ref{app:tab:dp-policy} and~\ref{app:tab:vla-policy} summarize
the training settings.

\subsection{Diffusion Policy}

We train the base diffusion policies on 75 demonstrations for
Pick-and-Place Strawberries and 100 each for Mug Hanging and Peg Insertion.
The A1 policy has seven proprioceptive/action coordinates (six arm joints
plus gripper); the Franka has eight (seven arm joints plus gripper). The wrist and external RGB
views each have their own group-normalized ResNet-18 encoder~\citep{wu2018groupnorm,kaimingresnet}, and image features are
fused with a proprioceptive MLP into a 512-dimensional feature per
observation step, giving 1024-dimensional conditioning for two steps. Images
are randomly cropped to $224\times224$ during training and center-cropped at
evaluation. The conditional 1D U-Net uses channel widths $[64,128]$, kernel
size~5, and a 128-dimensional diffusion-step embedding.

\begin{table}[!htbp]
\centering
\caption{Diffusion policy settings for A1 and Franka.}
\label{app:tab:dp-policy}
\footnotesize
\renewcommand{\arraystretch}{1.0}
\begin{tabular}{@{}l@{\hspace{5em}}r@{}}
\toprule
Setting & Value \\
\midrule
Initialization & ImageNet ResNet-18 \\
Trainable params. & $\approx$27.8M \\
Image input & 2 views \\
Image resolution & $224\times 224$ \\
Optimizer & AdamW, wd 0 \\
Batch size & 128 \\
Peak / min.\ LR & $7\times10^{-4}$ / $5\times10^{-6}$ \\
Warmup / cosine & 1k / 300k steps \\
Training steps & 200k \\
Grad.\ clipping & 1.0 \\
Precision / EMA & float32 / none \\
Prediction horizon & 16 \\
Execution horizon & 8 \\
Seed & 42 \\
\bottomrule
\end{tabular}
\end{table}

The learning-rate cosine schedule spans 300k steps, so training ends at
200k steps before reaching the minimum learning rate.
The model predicts noise in normalized action sequences with 100 diffusion
timesteps and a \appcode{squaredcos_cap_v2} schedule. Following Diffusion Policy~\citep{chi2023diffusion}, given a normalized
action sequence $\mathbf A$ noised at diffusion step $k$ to
$\mathbf A^{(k)}=\sqrt{\bar\alpha_k}\mathbf A+\sqrt{1-\bar\alpha_k}\epsilon$,
conditioning $z$, and a mask $m_u$ that is zero for episode-end padding
and one for valid actions, the objective is
\begin{equation}
 \mathcal L_{\mathrm{DP}}=
 \mathbb E\!\left[
 \frac{\sum_u m_u\|\epsilon_\theta(\mathbf A^{(k)},k,z)_u-
 \epsilon_u\|_2^2}{d_a\sum_u m_u}\right].
 \label{app:eq:dp-loss}
\end{equation}
Here, $u$ indexes actions within the predicted sequence, $d_a$ is the action
dimension, $\epsilon$ is standard Gaussian noise, and $\bar\alpha_k$ is the
cumulative product of the diffusion noise schedule's $\alpha$ coefficients
through step $k$.

At inference, 16 DDIM steps~\citep{song2021ddim} denoise a Gaussian action sequence. The first
predicted action is discarded to align the plan with the latest
observation, and eight actions (about 0.27\,s at 30\,Hz) are executed before
replanning. Takeover and episode reset clear the plan and observation
history (Appendix~\ref{app:takeover}).

\subsection{Vision--Language--Action Policy}

The $\pi_{0.5}$ configuration~\citep{intelligence2025pi05} uses a PaliGemma backbone
with the \appcode{gemma_2b} language-model configuration and the
\appcode{gemma_300m} action-expert configuration. The full model, including
the vision encoder and adaptive normalization, has approximately 3.35B
parameters.
It uses discrete state input. All model weights are fine-tuned with AdamW
using $(\beta_1,\beta_2)=(0.9,0.95)$.
Each VLA task starts from 100 base demonstrations. A 40k-step training run
takes approximately seven hours on a single $8\times$H100 node.

\begin{table}[!htbp]
\centering
\caption{$\pi_{0.5}$ settings for R1~Pro.}
\label{app:tab:vla-policy}
\footnotesize
\renewcommand{\arraystretch}{1.0}
\begin{tabular}{@{}l@{\hspace{5em}}r@{}}
\toprule
Setting & Value \\
\midrule
Initialization & $\pi_{0.5}$ base checkpoint \\
Trainable params. & $\approx$3.35B (full model) \\
Image input & 3 views \\
Observation horizon & 1 (current observation) \\
Image resolution & $224\times 224$ \\
Optimizer & AdamW, wd $10^{-10}$ \\
Batch size & 64 \\
Peak / min.\ LR & $2.5\times10^{-5}$ / $2.5\times10^{-6}$ \\
Warmup / cosine & 1k / 30k steps \\
Training steps & 40k \\
Grad.\ clipping & 1.0 \\
Precision / EMA & bfloat16 / 0.99 \\
Prediction horizon & 30 \\
Execution horizon & 15 \\
Seed & 42 \\
\bottomrule
\end{tabular}
\end{table}

The data pipeline combines three camera views, robot state,
actions, and the task prompt; resizes images to $224\times224$; tokenizes the
prompt (maximum 200 tokens); and pads state and action vectors to 32
dimensions. Torso and arm joint targets are converted to differences from
the current joint positions during training and restored to absolute
targets at deployment. Base velocity and gripper commands are excluded
from this conversion. For real R1~Pro, gripper commands use a continuous
$[-1,1]$ range, with $-1$ open and $+1$ closed.
States and action targets use mean--standard-deviation normalization;
deployment loads the normalization statistics saved with the checkpoint
and reverses action normalization before execution.

Table~\ref{app:tab:task-prompts} lists the task prompts supplied to the VLA
from the dataset metadata.

\begin{table*}[!t]
\centering
\caption{Language instructions used for the VLA tasks.}
\label{app:tab:task-prompts}
\footnotesize
\setlength{\tabcolsep}{6pt}
\renewcommand{\arraystretch}{1.12}
\begin{tabular}{@{}p{0.28\textwidth}p{0.68\textwidth}@{}}
\toprule
Task & Language instruction \\
\midrule
\raggedright Open Fridge and Pick Up Radio &
\raggedright Open the fridge, then pick up and turn on the radio. \tabularnewline
\raggedright Boxing Books up for Storage &
\raggedright Place the basket onto the table, and put the two books from the shelf into the basket. \tabularnewline
\raggedright Make Microwave Popcorn &
\raggedright Pick up the red popcorn bag from the shelf and place it into the microwave. \tabularnewline
\raggedright Prepare Make-Ahead Breakfast Bowls &
\raggedright Retrieve the bowl from the drawer, put the two bread in from the plate, and place the bowl on the right side of the countertop. \tabularnewline
\raggedright Turning on Radio &
\raggedright Pick up the radio on the table and poke the orange button. \tabularnewline
\bottomrule
\end{tabular}
\end{table*}

Deployment uses exponential-moving-average (EMA) model weights with a
decay of 0.99. The policy predicts 30 actions and executes 15 before
replanning.

\section{Evaluation Metrics and Experimental Protocols}
\label{app:evaluation}

\subsection{Autonomous Evaluation and Collection Outcomes}

Section~\ref{sec:setup} describes the tasks and platforms, and
Figure~\ref{fig:corrections} shows their annotated subtask sequences.
We summarize the task-success criteria below.

\paragraph{Simulation}
Pick-and-Place Strawberries requires both strawberries in the bowl.
Open Fridge and Pick Up Radio requires opening the refrigerator and
lifting the radio from the separate table.

\paragraph{Fixed-base manipulation}
Mug Hanging requires leaving the mug supported by its handle on the rack;
Peg Insertion requires inserting the peg into the matching fixture.

\paragraph{Whole-body mobile manipulation}
Prepare Make-Ahead Breakfast Bowls requires retrieving a bowl, closing the
drawer, and placing the bowl containing both bread items on the far side of
the countertop. Boxing Books up for Storage requires both books in the
basket on the table. Make Microwave Popcorn requires placing the popcorn inside the
microwave and closing its door. Turning on Radio requires picking up the
radio and pressing its button to turn it on while holding it.

\paragraph{Evaluation metrics}
Success rate is the fraction of autonomous evaluation episodes that
complete the full task. Table~\ref{tab:algorithm_comparison} reports mean success and sample standard
deviation across five evaluation seeds for Pick-and-Place Strawberries and
Open Fridge and Pick Up Radio, with 50 rollouts per seed.
The five seeds evaluate the same trained checkpoint under different
evaluation randomizations. Tables~\ref{tab:open_radio_seed_results} and~\ref{tab:pnp_seed_results}
provide the per-seed simulation results underlying
Table~\ref{tab:algorithm_comparison}, together with the base-policy and
additional-demonstration baselines. Real-world results report successes out of
25 evaluation episodes per condition.

\begin{table*}[!t]
    \centering
    \caption{Open Fridge and Pick Up Radio simulation success rates (\%) across five evaluation seeds. For correction-learning methods, parentheses indicate cumulative corrective trajectories.}
    \label{tab:open_radio_seed_results}
    \scriptsize
    \setlength{\tabcolsep}{4pt}
    \begin{tabular*}{\textwidth}{@{\extracolsep{\fill}}llrrrrrr@{}}
        \toprule
        Method & Setting & Seed 0 & Seed 1 & Seed 2 & Seed 3 & Seed 4 & Mean \\
        \midrule
        \multirow{2}{*}{Demo-only}
            & Base (100 demos)        & 36.0 & 48.0 & 38.0 & 36.0 & 40.0 & 39.6 \\
            & Additional demos (200 total) & 30.0 & 28.0 & 42.0 & 26.0 & 26.0 & 30.4 \\
        \midrule
        \multirow{5}{*}{Residual}
            & Single round (100) & 10.0 & 12.0 & 46.0 &  6.0 & 34.0 & 21.6 \\
            & Round 1 (25)       & 16.0 & 14.0 & 18.0 & 14.0 & 18.0 & 16.0 \\
            & Round 2 (50)       & 18.0 & 30.0 & 28.0 & 24.0 & 20.0 & 24.0 \\
            & Round 3 (75)       & 20.0 & 20.0 & 40.0 & 24.0 & 46.0 & 30.0 \\
            & Round 4 (100)      & 16.0 & 14.0 & 20.0 & 18.0 & 30.0 & 19.6 \\
        \midrule
        \multirow{5}{*}{BC (all data)}
            & Single round (100) & 86.0 & 84.0 & 88.0 & 56.0 & 82.0 & 79.2 \\
            & Round 1 (25)       & 72.0 & 78.0 & 66.0 & 68.0 & 70.0 & 70.8 \\
            & Round 2 (50)       & 84.0 & 88.0 & 92.0 & 86.0 & 82.0 & 86.4 \\
            & Round 3 (75)       & 82.0 & 94.0 & 94.0 & 86.0 & 80.0 & 87.2 \\
            & Round 4 (100)      & 84.0 & 88.0 & 94.0 & 86.0 & 88.0 & 88.0 \\
        \midrule
        \multirow{5}{*}{SIRIUS}
            & Single round (100) & 70.0 & 86.0 & 82.0 & 54.0 & 74.0 & 73.2 \\
            & Round 1 (25)       & 76.0 & 36.0 & 48.0 & 78.0 & 36.0 & 54.8 \\
            & Round 2 (50)       & 58.0 & 66.0 & 56.0 & 74.0 & 88.0 & 68.4 \\
            & Round 3 (75)       & 82.0 & 50.0 & 84.0 & 54.0 & 82.0 & 70.4 \\
            & Round 4 (100)      & 78.0 & 66.0 & 64.0 & 68.0 & 82.0 & 71.6 \\
        \midrule
        \multirow{5}{*}{HG-DAgger}
            & Single round (100) & 86.0 & 90.0 & 82.0 & 54.0 & 88.0 & 80.0 \\
            & Round 1 (25)       & 70.0 & 58.0 & 40.0 & 44.0 & 46.0 & 51.6 \\
            & Round 2 (50)       & 70.0 & 80.0 & 66.0 & 68.0 & 78.0 & 72.4 \\
            & Round 3 (75)       & 84.0 & 90.0 & 88.0 & 90.0 & 58.0 & 82.0 \\
            & Round 4 (100)      & 88.0 & 88.0 & 90.0 & 90.0 & 94.0 & \textbf{90.0} \\
        \bottomrule
    \end{tabular*}
\end{table*}

\begin{table*}[!t]
    \centering
    \caption{Pick-and-Place Strawberries simulation success rates (\%) across five evaluation seeds. For correction-learning methods, parentheses indicate cumulative corrective trajectories.}
    \label{tab:pnp_seed_results}
    \scriptsize
    \setlength{\tabcolsep}{4pt}
    \begin{tabular*}{\textwidth}{@{\extracolsep{\fill}}llrrrrrr@{}}
        \toprule
        Method & Setting & Seed 0 & Seed 1 & Seed 2 & Seed 3 & Seed 4 & Mean \\
        \midrule
        \multirow{2}{*}{Demo-only}
            & Base (75 demos)              & 50.0 & 42.0 & 58.0 & 48.0 & 68.0 & 53.2 \\
            & Additional demos (175 total) & 52.0 & 54.0 & 62.0 & 54.0 & 56.0 & 55.6 \\
        \midrule
        \multirow{5}{*}{Residual}
            & Single round (100) & 48.0 & 44.0 & 48.0 & 42.0 & 44.0 & 45.2 \\
            & Round 1 (25)       & 56.0 & 46.0 & 50.0 & 46.0 & 48.0 & 49.2 \\
            & Round 2 (50)       & 48.0 & 46.0 & 46.0 & 44.0 & 44.0 & 45.6 \\
            & Round 3 (75)       & 50.0 & 36.0 & 44.0 & 30.0 & 40.0 & 40.0 \\
            & Round 4 (100)      & 56.0 & 42.0 & 42.0 & 36.0 & 58.0 & 46.8 \\
        \midrule
        \multirow{5}{*}{BC (all data)}
            & Single round (100) & 12.0 & 14.0 & 16.0 & 14.0 & 18.0 & 14.8 \\
            & Round 1 (25)       &  0.0 &  0.0 &  0.0 &  2.0 &  0.0 &  0.4 \\
            & Round 2 (50)       &  0.0 &  2.0 &  2.0 &  0.0 &  2.0 &  1.2 \\
            & Round 3 (75)       &  8.0 &  2.0 &  2.0 &  6.0 &  4.0 &  4.4 \\
            & Round 4 (100)      & 10.0 &  6.0 & 16.0 &  6.0 & 16.0 & 10.8 \\
        \midrule
        \multirow{5}{*}{SIRIUS}
            & Single round (100) & 76.0 & 78.0 & 74.0 & 74.0 & 78.0 & 76.0 \\
            & Round 1 (25)       & 84.0 & 64.0 & 64.0 & 76.0 & 66.0 & 70.8 \\
            & Round 2 (50)       & 76.0 & 86.0 & 76.0 & 76.0 & 80.0 & 78.8 \\
            & Round 3 (75)       & 46.0 & 28.0 & 40.0 & 34.0 & 40.0 & 37.6 \\
            & Round 4 (100)      & 80.0 & 74.0 & 74.0 & 78.0 & 88.0 & 78.8 \\
        \midrule
        \multirow{5}{*}{HG-DAgger}
            & Single round (100) & 90.0 & 90.0 & 92.0 & 90.0 & 84.0 & 89.2 \\
            & Round 1 (25)       & 78.0 & 84.0 & 74.0 & 76.0 & 66.0 & 75.6 \\
            & Round 2 (50)       & 86.0 & 90.0 & 84.0 & 94.0 & 88.0 & 88.4 \\
            & Round 3 (75)       & 92.0 & 94.0 & 82.0 & 86.0 & 88.0 & 88.4 \\
            & Round 4 (100)      & 96.0 & 90.0 & 94.0 & 94.0 & 90.0 & \textbf{92.8} \\
        \bottomrule
    \end{tabular*}
\end{table*}

\subsection{Intervention Burden and Aggregation}

Figure~\ref{fig:intervention_trends} uses the annotations and takeover
filtering described in Appendix~\ref{app:subtasks}. Within the union of
annotated intervals, let $\widetilde h_{e,t}$ indicate a retained intervention
frame (zero elsewhere),
$T_e$ the episode length, and $M_e$ the number of frames covered by the
annotations. The takeover count and intervention fraction are
\begin{equation}
 \begin{split}
 K_e&=\widetilde h_{e,0}+\sum_{t=1}^{T_e-1}
 (1-\widetilde h_{e,t-1})\widetilde h_{e,t},\\
 F_e&=\frac{1}{M_e}\sum_{t=0}^{T_e-1}\widetilde h_{e,t}.
 \end{split}
 \label{app:eq:burden}
\end{equation}
The plotted percentage is $100F_e$. Takeover duration is computed as
$D_e=\sum_t\widetilde h_{e,t}/f_e$, where $f_e$ is the frame rate stored
in the annotation metadata. Each metric is averaged across episodes within
a task and round; the figure's overall mean gives each task equal weight.

For the operator study, cumulative intervention share pools recorded
timesteps across episodes through round $r$, separately for each operator:
\begin{equation}
 F^{\mathrm{cum}}_r=
 \frac{\sum_{j=1}^{r}\sum_e\sum_t h_{j,e,t}}
 {\sum_{j=1}^{r}\sum_e T_{j,e}}.
 \label{app:eq:cumulative}
\end{equation}
Here, $h_{j,e,t}$ is the recorded intervention indicator and $T_{j,e}$ is
the episode length. Cumulative takeover-free episode counts sum episodes
with no recorded intervention through the same round.

\subsection{Whole-Body Activation Analysis}

The body-part analysis pools four 25-episode rounds. A body part is active
when any of its coordinates exceeds its activity threshold in magnitude
for three consecutive frames. Base activity uses the commanded velocities,
with thresholds of 0.003\,m/s for translation and 0.003\,rad/s for rotation.
Torso, arm, and gripper activity uses absolute action--state differences,
with thresholds of 0.01\,rad, 0.02\,rad, and 0.75, respectively; the gripper
difference is in the normalized command range. Gripper activity is assigned
to the corresponding arm. Each body part receives at most one vote per
takeover segment containing qualifying activity.
If $v_{kb}\in\{0,1\}$ denotes the vote of segment $k$ for body part $b$,
the normalized activation share is
\begin{equation}
 V_b=\frac{\sum_k v_{kb}}{\sum_{b'}\sum_k v_{kb'}}.
 \label{app:eq:votes}
\end{equation}
Multiple body parts can contribute during one takeover. Each part's share
is its activation count divided by the total across all body parts.

\subsection{Baseline Configurations}

The demonstration-only and correction-based conditions each add 100
trajectories to the same task-specific base demonstration set. The
collection-schedule comparison uses either 100 correction episodes
collected against the fixed base policy or four rounds of 25 episodes
with policy updates between rounds.

The algorithm study compares HG-DAgger~\citep{kelly2019hgdagger}, all-data BC,
SIRIUS~\citep{liu2023sirius}, and residual learning. Each method collects
corrections with its own policy and trains on its accumulated data.
Appendix~\ref{app:processing} specifies the sampling rules for HG-DAgger,
BC, and SIRIUS.

\paragraph{Residual learning}
The residual policy adds a correction to a frozen base policy $\pi_0$
(DP for A1, $\pi_{0.5}$ for R1~Pro). Each camera is encoded by an
ImageNet-initialized ResNet-18, and the image features, proprioception, and
the normalized base action chunk are fused and decoded by an MLP with a
zero-initialized output layer, so the initial residual is zero. The
residual chunk has the same length as the base policy's execution horizon,
8 steps for A1 and 15 steps for R1~Pro. It is queried whenever the base
replans, and step $k$ of the residual chunk is added to step $k$ of the
base chunk. There is no intervention head, and the correction is always
applied.

At every step of collection we store the base policy's predicted action
$\mathbf a^{\pi_0}_t$ alongside the executed action
$\mathbf a^{\mathrm{exec}}_t$, both in the normalized joint space of
Eq.~\eqref{app:eq:normalization}. The residual target is
\begin{equation}
 \Delta\mathbf a_t=\mathbf a^{\mathrm{exec}}_t-\mathbf a^{\pi_0}_t,
 \label{app:eq:residual}
\end{equation}
where arm, torso, and base channels are clipped to $[-2,2]$ and halved, so
the network regresses a target in $[-1,1]$ following TRANSIC~\citep{jiang2024transic}.
At deployment the network output is multiplied by two before it is added to
the base chunk, and the composed command is clipped to $[-1,1]$. Grippers are
predicted as absolute $\{-1,+1\}$ commands rather than residuals, following
the human during corrections and the base policy otherwise. Correction
frames (Appendix~\ref{app:labels}) use the target above, demonstration and
autonomous frames use a zero target, and pre-intervention frames are
excluded. The loss is the mean squared error over valid chunk steps, with
an equally weighted penalty pulling the residual to zero on non-correction
frames. Chunk starts are sampled with probability 0.3 from demonstrations,
0.5 from corrections, and 0.2 from autonomous frames, and one fifth of the
correction draws come from the first 20 correction frames of each episode,
where the residual has to diverge from the base. We train with AdamW at
learning rate $10^{-4}$ with 1{,}000 warmup steps and cosine decay to
$10^{-6}$, batch size 128, for 200k steps.

Unlike the other methods, the base policy is never retrained. In round $k$,
the robot runs the base policy together with the round-$(k{-}1)$ residual
while corrections are collected. The stored base action is still the frozen
base policy's prediction, so the target in Eq.~\eqref{app:eq:residual} has
the same meaning in every round. After each round, a new residual is
trained from initialization on the demonstrations plus all correction
rounds collected so far. Demonstrations store the executed action as the
base action, which gives them a zero target.

\subsection{Operator Study Protocol}

The operator study includes three participants recruited from the lab
(two male and one female).
Two participants had prior experience with robot teleoperation, and one was a novice.
Before the study's correction collection began, each participant completed
approximately 25 familiarization episodes.
Participants were instructed to intervene when failure seemed likely and
resume autonomous control once the robot could continue. All three started
from the same base-policy checkpoint, then collected corrections and
retrained their policies separately.
Each participant collected three rounds of 25 retained episodes on Franka
Mug Hanging, for 225 episodes in total. Failed attempts were excluded from
the dataset. Figure~\ref{fig:operator-study} computes intervention metrics
over these retained episodes and identifies takeover-free episodes by the
absence of recorded interventions.

\FloatBarrier
\end{document}